\documentclass[sigconf]{acmart}

\AtBeginDocument{%
  \providecommand\BibTeX{{%
    \normalfont B\kern-0.5em{\scshape i\kern-0.25em b}\kern-0.8em\TeX}}}

\setcopyright{acmcopyright}
\copyrightyear{2023} 
\acmYear{2023}  
\setcopyright{acmlicensed}
\acmConference[MM '23]{Proceedings of the 31st ACM International Conference on Multimedia}{October 29-November  3, 2023}{Ottawa, ON, Canada}
\acmBooktitle{Proceedings of the 31st ACM International Conference on Multimedia (MM '23), October   29-November  3, 2023, Ottawa, ON, Canada}
\acmPrice{15.00}
\acmDOI{10.1145/3581783.3612413}
\acmISBN{979-8-4007-0108-5/23/10}

\usepackage{enumitem}
\setlist[enumerate]{listparindent=\parindent}
\begin{document}

\title{Towards Better Multi-modal Keyphrase Generation via Visual Entity Enhancement and Multi-granularity Image Noise Filtering}

\author{Yifan Dong}
\authornote{Both authors contributed equally to this research.}
\email{dongyifan@stu.xmu.edu.cn}
\affiliation{%
  \institution{Department of Software Engineering, Xiamen University}
  \country{ }
  }

\author{Suhang Wu}
\authornotemark[1]
\email{wush_xmu@outlook.com}
\affiliation{%
  \institution{Department of Digital Media Technology, Xiamen University}
\country{ }
  }

\author{Fandong Meng}
\email{fandongmeng@tencent.com}
\affiliation{%
  \institution{Tencent WeChat AI}
\country{ }
  }

\author{Jie Zhou}
\email{withtomzhou@tencent.com}
\affiliation{%
\country{ }
  \institution{Tencent WeChat AI}
  }
  
\author{Xiaoli Wang}
\email{xlwang@xmu.edu.cn}
\affiliation{%
  \institution{Department of Software Engineering, Xiamen University}\country{}
  }

\author{Jianxin Lin}
\email{linjianxin@hnu.edu.cn}
\affiliation{%
  \institution{Hunan University}\country{}
  }
  
\author{Jinsong Su}
\authornote{Corresponding Author.}
\email{jssu@xmu.edu.cn}
\affiliation{%
  \institution{Department of Artificial Intelligence, Xiamen University}
   \country{}
  }

\renewcommand{\shortauthors}{Yifan Dong, et al.}
\renewcommand{\shorttitle}{Towards Better Multi-modal Keyphrase Generation via Visual
Entity Enhancement and Image Noise Filtering} 

\begin{abstract}
\begin{sloppypar}
Multi-modal keyphrase generation aims to produce a set of keyphrases that represent the core points of the input text-image pair. 
In this regard, dominant methods mainly focus on multi-modal fusion for keyphrase generation. Nevertheless, there are still two main drawbacks: 
1) only a limited number of sources, such as image captions, can be utilized to provide auxiliary information. However, they may not be sufficient for the subsequent keyphrase generation.
2) the input text and image are often not perfectly matched, and thus the image may introduce noise into the model.
To address these limitations, in this paper, we propose a novel multi-modal keyphrase generation model, which not only enriches the model input with external knowledge, but also effectively filters image noise. First, we introduce external visual entities of the image as the supplementary input to the model,
which benefits the cross-modal semantic alignment for keyphrase generation. Second, we simultaneously calculate an image-text matching score and image region-text correlation scores to perform multi-granularity image noise filtering. Particularly, we introduce the correlation scores between image regions and ground-truth keyphrases to refine the calculation of the previously-mentioned correlation scores. To demonstrate the effectiveness of our model, we conduct several groups of experiments on the benchmark dataset.
 Experimental results and in-depth analyses show that our model achieves the state-of-the-art performance. Our code is available on https://github.com/DeepLearnXMU/MM-MKP.
\end{sloppypar}
\end{abstract}





\maketitle

\section{Introduction}
\begin{figure}[t]
  \centering
  \includegraphics[width=\linewidth]{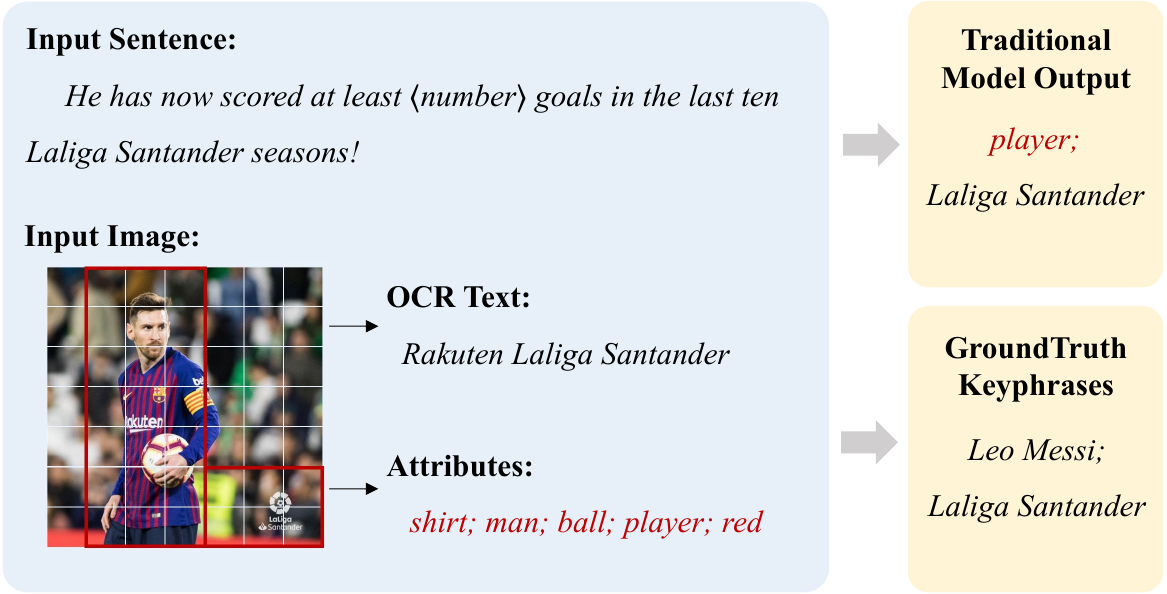}
  \caption{An example of multi-modal keyphrase generation. We can observe that the image attributes are both coarse and limited. Moreover, certain image regions may be irrelevant to the sentence, consequently leading to the introduction of noise into the model.}
  \Description{}
  \label{fig:1}
\end{figure}

With the rapid development of social networking platforms, humans often express their viewpoints and emotions with multi-modal information, which may contain both images and texts. Hence, multi-modal keyphrase generation has become an emerging task, which aims to output keyphrases with the given input text and image. Figure \ref{fig:1} provides an example of this task. Given the input text-image pair, we extract textual and visual features, and then fuse these features to generate keyphrases. Compared to traditional text-only keyphrase generation \cite{DBLP:conf/acl/MengZHHBC17, DBLP:conf/acl/ChenCLK20, DBLP:conf/acl/YeGL0Z20, DBLP:conf/emnlp/XieWYLXWZS22}, multi-modal keyphrase generation focuses on exploiting the complementarity of image and text to generate better keyphrases. It not only facilitates the understanding of how humans utilize multi-modal information to comprehend and summarize the input text, but also holds broad applicability in various real-life scenarios, including opinion mining and content recommendation. Therefore, multi-modal keyphrase generation has attracted increasing attention in recent years.

In the line of exploring this task, early studies hold that hashtags can express important information in multi-media content, and thus directly treat hashtags as keyphrases \cite{DBLP:conf/ijcai/GongZ16, DBLP:conf/sigir/SedhaiS14, DBLP:conf/ijcai/ZhangWHHG17, DBLP:conf/aaai/Zhang00TY019}. Therefore, multi-modal keyphrase generation is usually modeled as a hashtag recommendation task. Typically, these studies adopt a co-attention network to fuse textual and visual tweet information for recommending hashtags \cite{DBLP:conf/ijcai/ZhangWHHG17, DBLP:conf/aaai/Zhang00TY019}. Unlike the studies mentioned above, \citet{DBLP:conf/emnlp/WangLLK20} first perform Optical Character Recognition (OCR) to extract explicit optical characters from the input image and then utilize an image captioning model to extract implicit image attributes that reflect the semantic information of the image. To better integrate multi-modal information, they then introduce a multi-modal multi-head attention to model the semantic interactions between different modalities. Besides, during the model training, they introduce a pointer network \cite{DBLP:conf/acl/GuLLL16} to output keyphrases, where keyphrase classification and generation are jointly modeled.
Particularly, this model achieves state-of-the-art (SOTA) performance on the commonly-used multi-modal keyphrase dataset. 

In spite of their success, there are still some defects in the above studies. First, these studies often represent the semantic information of each input image with attributes. Usually, these attributes are nouns and adjectives extracted from the caption of input image. However, attributes may be coarse and limited for this task. As shown in Figure \ref{fig:1}, we can observe that the extracted image attributes include ``\emph{man}'' which merely denotes the object as a person without providing any further elaboration. Besides, by analyzing the attributes extracted from training data, we observe that more general concepts such as color, shape, and person constitute the majority of attributes, which cannot provide effective supplementary information for keyphrase generation. 
Second, the aforementioned studies place greater emphasis on multi-modal fusion while neglecting the possible mismatch between text and image in social multi-media data. To study its effect, we use the commonly-used pre-trained multi-modal model ViLT \cite{DBLP:conf/icml/KimSK21} to obtain image and text representations and then conduct text-image matching analysis. Our findings reveal that only 63\% of the text-image pairs can achieve the matching score exceeding 0.8. Even for relevant text-image pairs, there often exist some image regions that do not match the text. As shown in Figure \ref{fig:1}, the regions marked by red boxes are highly relevant to the text, while other regions are less relevant to the text. Such regions are likely to introduce noise to the model and degrade the model performance consequently. Therefore, how to effectively exploit input images remains still a challenge for multi-modal keyphrase generation.

\begin{figure*}[h]
  \centering
  \includegraphics[width=\linewidth]{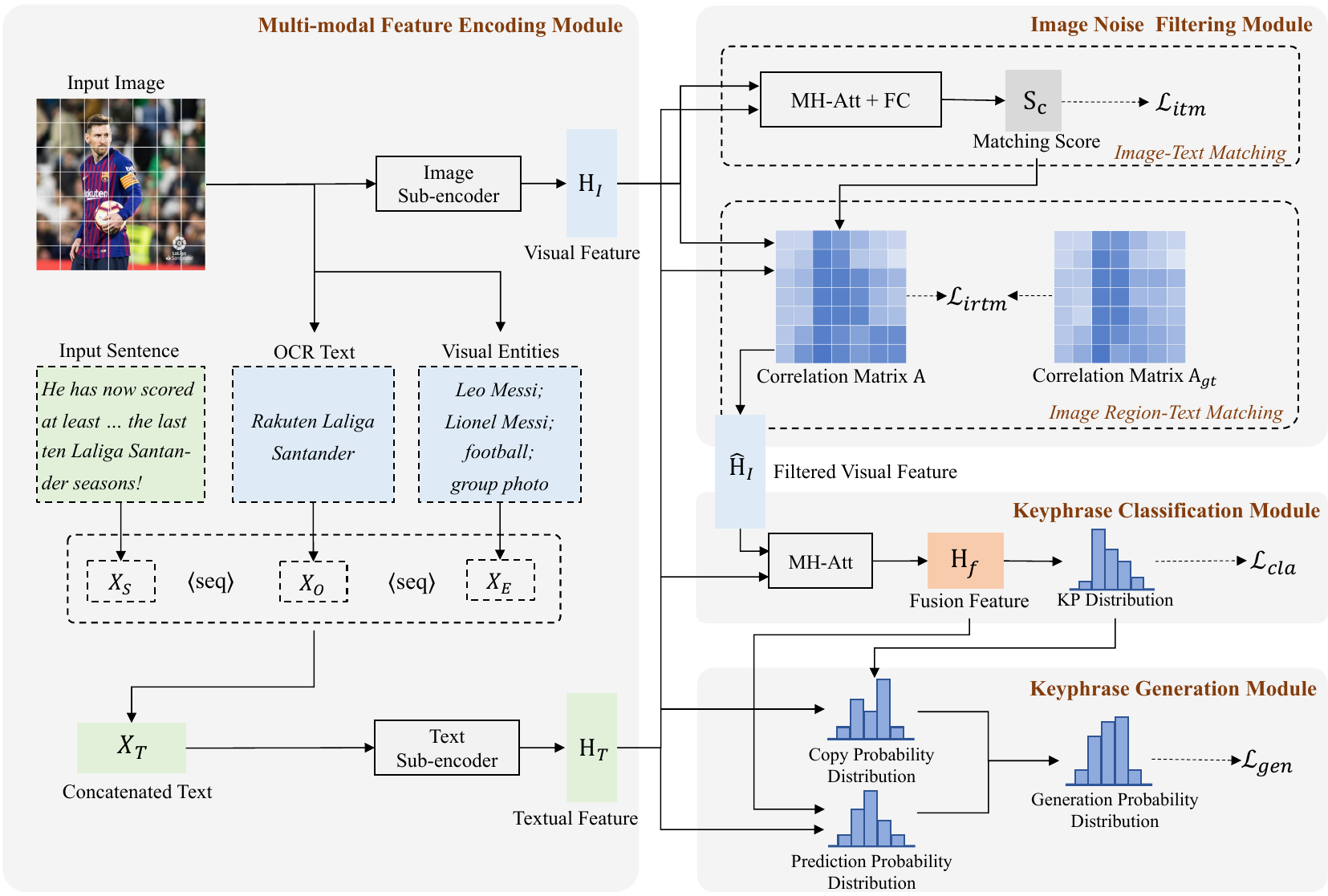}
  \caption{The overall architecture of our model, including multi-modal feature encoding module, image noise filtering module, keyphrase classification module and keyphrase generation module.  }
  \Description{}
  \label{fig:model}
\end{figure*}

In this work, we propose a novel multi-modal keyphrase generation model with visual entity enhancement and image noise filtering. Our model is a significant extension of \cite{DBLP:conf/emnlp/WangLLK20}. Our model not only introduces external visual entities as supplementary information of the textual input, but also leverages multi-granularity noise filtering to effectively exploit the input image. As shown in Figure \ref{fig:model}, our model mainly contains four modules: \textbf{1) Multi-modal semantic encoding module.} This module comprises two sub-encoders that learn the semantic representations of input text and image respectively. 
When encoding the input text, we first use Baidu API\footnote{https://ai.baidu.com/tech/imagerecognition} to acquire the visual entities semantically related to the input image. Then, the text sub-encoder receives a concatenation of the OCR text, original input text, and the acquired visual entities. To distinguish their effects, type embeddings are introduced. \textbf{2) Image noise filtering module.} In this module, we explore two strategies to perform multi-granularity image noise filtering. One is image-text matching. Adopting this strategy, we calculate a matching score between the input text and the whole image. The other is image region-text matching. Using this strategy, we divide the whole input image into $7\times7$ regions, and calculate the correlation score between each image region and the text, forming a correlation matrix. Then, the region-level vector representations of the input image are weighted with the matching score and correlation matrix. \textbf{3) Keyphrase classification module.} It fuses the filtered image and text representations to perform keyphrase classification. \textbf{4) Keyphrase generation module.} This module is based on a pointer network, where both the concatenated text input and keyphrase classification are exploited to generate each keyphrase as a sequence. 

To train our model, in addition to the conventional keyphrase generation loss and keyphrase classification loss, we introduce two additional losses. The first one is an image-text matching loss, to identify whether the image and text are relevant. The second one is an image region-text correlation score divergence loss. Particularly, we utilize the correlation matrix between ground-truth keyphrases and image regions as the supervisory information, allowing the model to automatically concentrate on image regions that are useful for keyphrase generation.

Compared with the previous models \cite{DBLP:conf/ijcai/ZhangWHHG17, DBLP:conf/aaai/Zhang00TY019, DBLP:conf/emnlp/WangLLK20}, our model possesses the following two advantages. First, we introduce visual entities that are semantically related to the input image and can be served as anchor points for cross-modal semantic alignment. Second, we utilize multi-granularity noise filtering strategies to more effectively exploit the input image.

To investigate the effectiveness of our model, we conduct several groups of experiments on the benchmark dataset. Experimental results and in-depth analyses show that our model outperforms the current best model, achieving the SOTA performance.

\section{RELATED WORK}

\textbf{Keyphrase Generation.} The task of keyphrase generation has received sustained attention in recent years. The commonly-used models for keyphrase generation can be roughly classified into extraction and generation approaches. Early studies mainly focus on using statistical models to perform keyphrase extraction \cite{DBLP:journals/ipm/SaltonB88, DBLP:journals/is/El-BeltagyR09, DBLP:conf/ecir/0001MPJNJ18a, DBLP:journals/ipm/XieSSWWYLXS23}. With the rapid development of deep learning, a number of neural network based models have been proposed for keyphrase generation. Generally, the frameworks for keyphrase generation can be divided into three categories: 1) One2one \cite{DBLP:conf/emnlp/ChenZ0YL18, DBLP:conf/aaai/ChenGZKL19, DBLP:conf/acl/MengZHHBC17}. This category splits a training instance into multiple pairs, each consisting of the input text and only one corresponding keyphrase. During inference, it adopts beam search to produce candidate phrases and then selects the top-$K$ ranked ones as the final keyphrases. 2) One2seq \cite{DBLP:conf/acl/YuanWMTBHT20, DBLP:conf/acl/ChenCLK20}, which concatenates all keyphrases in a given order as a training instance. During inference, the model outputs all keyphrases as a sequence. 3) One2set \cite{DBLP:conf/acl/YeGL0Z20}. In this category, the generation of keyphrases is modeled as a generation task of a keyphrase set, where keyphrases are individually generated in parallel. Unlike these studies focusing on text-only keyphrase generation, we set our sights on multi-modal keyphrase generation. In this aspect, the common practice \cite{DBLP:conf/ijcai/ZhangWHHG17, DBLP:conf/aaai/Zhang00TY019} use a co-attention network to fuse textual and visual information and then recommend tags for multi-modal tweets. \citet{DBLP:conf/emnlp/WangLLK20} propose a multi-modal keyphrase generation model based on an encoder-decoder framework. Typically, the encoder is equipped with a multi-head attention mechanism to fuse multi-modal information, and the decoder is a pointer network.

\textbf{Multi-modal Fusion}. How to effectively fuse multi-modal information is always a hot research topic. Dominant approaches can be roughly classified into the following three categories \cite{DBLP:journals/jstsp/ZhangYHD20}: 1) simple operations such as concatenation \cite{DBLP:journals/corr/abs-1903-02930}, weighted sum with scalar weights \cite{DBLP:conf/cvpr/Perez-RuaVPBJ19} and progressive exploration decision fusion  \cite{DBLP:journals/corr/abs-1712-00559, DBLP:conf/bmvc/Perez-RuaBP18}; 2) bilinear pooling \cite{DBLP:conf/nips/KimJZ18, DBLP:conf/aaai/Ben-younesCTC19}; 3) attention-based methods, such as graph attention mechanisms \cite{DBLP:conf/naacl/JoshiBJSM22}, symmetric attention mechanisms \cite{DBLP:conf/ner/ZhaoLL21}, dual attention networks \cite{DBLP:conf/cvpr/NamHK17}, dynamic gated aggregation mechanisms \cite{DBLP:conf/naacl/ChenZLYDTHSC22}, and dynamic parameter prediction networks \cite{DBLP:conf/cvpr/NohSH16}.

Particularly, some studies concentrate on multi-modal fusion in the presence of image noise. For example, \citet{DBLP:conf/coling/SunWSWSZC20} present a pre-trained multi-modal model based on relationship inference and visual attention. Typically, it contains a gated unit that adjusts the weights of visual features during fusion based on the image-text matching score. \citet{DBLP:conf/aaai/0006W0SW21} propose the text-image relationship propagation to reduce the distraction of irrelevant images on the multi-modal named entity recognition task. \citet{DBLP:conf/ijcai/YuWXL22} put forward a coarse-to-fine image-target matching model for the target-oriented (aspect-based) multi-modal sentiment classification task. With extra manually labeled data, they explore two supervised tasks to capture the image-target matching relations for multi-modal fusion.
\citet{DBLP:conf/coling/YeGXT022} construct a cross-modal relation-aware attention module, which is equipped with a mask matrix based on the relevance of text and image regions. This matrix conducts noise filtering during the self-attention process, improving the performance of multi-modal machine translation.

To the best of our knowledge, our work is the first attempt to introduce external visual entities to improve multi-modal keyphrase generation. These entities not only provide supplementary information but also benefit cross-modal semantic alignment. Besides, we further explore multi-granularity noise filtering to exploit the input image more precisely. The subsequent experimental results strongly verify the effectiveness of visual entities and image noise filtering strategies.

\section{Our Model}

Before elaborating on our model, we first briefly introduce the formulation of this task. Given a text-image pair $(X_S, X_I)$ of the dataset $D$, multi-modal keyphrase generation aims to predict a keyphrase set $\mathcal{Y}$. Following \cite{DBLP:conf/acl/MengZHHBC17}, we replicate the original input pair multiple times to ensure that each input pair is associated with one keyphrase, forming a triplet set $\{(X_S, X_I, y)\}$, where $y\in{\mathcal{Y}}$.

In the subsequent subsections, we first give a description of the architecture of our model, and then describe details of the model training.

\subsection{Model Architecture}
Figure \ref{fig:model} illustrates the basic architecture of our model. Overall, our model includes four modules: 1) \emph{Multi-modal feature encoding module} learning the representations of the input text and image, respectively; 2) \emph{Image noise filtering module} that conducts multi-granularity image noise filtering to generate a better image representation; 3) \emph{Keyphrase classification module} that fuses the filtered image and text representations, and then performs keyphrase classification; 4) \emph{Keyphrase generation module}, which is based on a pointer network and generates each keyphrase in the form of a sequence. These modules are described in detail in the following.

\subsubsection{\textbf{Multi-modal Feature Encoding Module}} \

This module contains an image sub-encoder and a text sub-encoder, extracting visual features and textual features respectively. To provide this module with more information for better keyphrase generation, we first preprocess the input image to get the textual information contained in the image, including OCR and visual entity extraction. 

Specifically, we use the commonly-used PaddleOCR\footnote{https://github.com/PaddlePaddle/PaddleOCR} to extract the explicit optical characters (e.g., slogans) from the image. 
Meanwhile, we use the Baidu API\footnote{https://ai.baidu.com/tech/imagerecognition} to obtain highly-confident visual entities that are semantically related to the input image. 
As the example shown in Figure \ref{fig:model}, compared with the commonly-used attributes or image captions, these visual entities not only provide additional detailed textual descriptions of image objects, but also can serve as semantic anchors to facilitate cross-modal semantic alignment, thus leading to better keyphrase generations. To facilitate the subsequent descriptions, we denote the extracted OCR text and visual entities as $X_O$ and $X_E$, respectively.

Then, the original input text, OCR text, and visual entities are sequentially concatenated and fed to the text sub-encoder. Meanwhile, the input image is encoded by the image sub-encoder. 

{\bfseries{Text sub-encoder.}} To distinguish $X_O$ and $X_E$ from the original input text $X_S$, we insert two delimited $\langle\text{seq}\rangle$ tokens to respectively indicate the beginning positions of $X_O$ and $X_E$, obtaining the concatenated input of text modality: ${X_T}=X_S\langle\text{seq}\rangle X_O\langle\text{seq}\rangle X_E$.
Then, we feed $X_T$ into the text sub-encoder, which is based on Bi-GRU\footnote{We also try to use Transformer \cite{DBLP:conf/nips/VaswaniSPUJGKP17} as the fundamental architecture of our model. However, regardless of the setting used, our GRU-based multi-modal keyphrase generation model performs better than the Transformer-based one. Therefore, we follow our most important baseline \cite{DBLP:conf/emnlp/WangLLK20} and use GRU to build encoder and decoder.}, learning the token-level semantic representations of $X_T$:
\begin{equation}
    \text{H}_T = \text{Bi-GRU}(X_{emb}),
\end{equation}
where ${\text{H}_T}$$\in$${\mathbb{R}^{|X_T|\times d_1}}$, $d_1$ denotes the hidden state dimension, and $X_{emb}$ is the embedding sequence of $X_T$. Here we use the sum of word embedding and type embedding to represent each token. Besides, we use the pre-trained Glove \cite{DBLP:conf/emnlp/PenningtonSM14} word embedding to initialize the input word embedding, and randomly initialize the type embedding.

Finally, we obtain a global vector representation of text modality via max-pooling operation: $\text{M}_T=\text{Max-pooling}(\text{H}_T)$.

{\bfseries{Image sub-encoder.}} 
Following common practice \cite{DBLP:conf/coling/SunWSWSZC20}, we employ the pre-trained model VGG19 \cite{DBLP:journals/corr/SimonyanZ14a} to extract the visual features of each input image. Concretely, we first resize each image to $224\times224$ pixels and feed it to the VGG19 model. The last-layer output is a $7\times7\times512$-dimensional vector containing 49 local spatial region features for each image. That is, the visual feature of each region is represented as a 512-dimensional vector. To further use these visual features, we perform flattening and linear projection on these visual features:
\begin{equation}
    \vspace{-0.5ex}
    \text{H}_I=\text{flatten}(\text{VGG19}(X_I))W_I+b_I,
    \vspace{-0.5ex}
\end{equation}
where ${\text{H}_I}\in{\mathbb{R}^{49\times{d_1}}}$ and $\text{flatten}(\cdot)$ is a function reshaping the $7\times7\times512$-dimensional vector to a $49\times512$-dimensional one. Additionally, $W_I\in{\mathbb{R}^{49\times{d_1}}}$ and $b_I\in{\mathbb{R}^{49\times{d_1}}}$ are learnable parameter matrices. 

\subsubsection{\textbf{Image Noise Filtering Module}} \

In this module, we explore two cross-modal matching strategies to filter the noise of each input image, obtaining a filtered image representation. Via the combined effect of the two cross-modal matching strategies, this module may help the model focus on key regions for keyphrase generation while avoiding the interference of image noise.

{\bfseries{Image-text matching.}} \
Using this strategy, we obtain a score indicating the semantic matching degree between the whole image and the input text. Specifically, we first use a multi-head cross-attention function to the fusion representation $\text{H}_c$:
\begin{equation}
    \text{H}_c = \text{MultiHead}(\text{M}_T, \text{H}_I, \text{H}_I),
\end{equation}
where MultiHead(*) is a multi-head cross-attention function, the global textual feature $\text{M}_T$ is used as the query, and the visual feature $\text{H}_I$ works as the key and value.

On the top of $\text{H}_c$, we stack a fully-connected (FC) layer to perform image-text matching, where a matching score $s_c$ is acquired and then used in conjunction with the subsequent image region-text matching strategy to filter image noise.

{\bfseries{Image region-text matching}} \
This strategy is used to filter the irrelevant regions of the input image. 
To achieve this, we first project the visual feature $\mathrm{H}_{I}$ and the global textual feature $\mathrm{M}_{T}$ into a shared semantic space: $\overline{\mathrm{H}}_T=W_{T}\mathrm{M}_{T}$, $\overline{\mathrm{H}}_{I}=W_{I}\mathrm{H}_{I}$, facilitating the subsequent calculation of their semantic correlation. Here, $\mathrm{H}_I$ represents the flattened representation of $7\times7$ image region features, while $W_*$ are learnable parameter matrices. 

Subsequently, we calculate the image region-text correlation matrix $\mathrm{A}$ as fellow:
\begin{equation}
    \mathrm{A}= \text{FFN}\left(\frac{(\overline{\mathrm{H}}_T)\cdot({\overline{\mathrm{H}}_{I}})^\top}{\sqrt{d_2}}+J \times s_c \right), \label{eq:matrix}
\end{equation}
where FFN(*) is a feedforward network, $d_2$ is the dimension of vector representation in the shared semantic space, the element $\mathrm{A}_{l_1,l_2}$ indicates the semantic matching score between the input text and the $(7$$\times$$l_1$$+$$l_2)$-th image region, and $J$ is an all-ones matrix. Note that we use the above-mentioned image-text matching score $s_c$ to smooth the matrix $\mathrm{A}$.

Lastly, we use a Sigmoid function to produce a filtered image representation $\widehat{\mathrm{H}}_{I}$:
\begin{equation}
    \widehat{\mathrm{H}}_{I}=\text{Sigmoid}(\mathrm{A})\odot \mathrm{H}_{I}, \label{sig}
\end{equation}
where $\odot$ is the element-wise multiplication.

\subsubsection{\textbf{Keyphrase Classification Module}} \

Following \cite{DBLP:conf/emnlp/WangLLK20}, we also regard each keyphrase in training data as a discrete label and directly use a classifier to predict keyphrases.

Concretely, we first use a multi-head cross-attention to effectively fuse the filtered visual and textual features, and then use an FFN with residual connection and layer normalization to obtain a fused vector $\mathrm{H}_{f}$:
\begin{equation}
    \mathrm{H}_{f} = \text{FFN}(\text{MultiHead}(\mathrm{M}_T, \widehat{\mathrm{H}}_{I},\widehat{\mathrm{H}}_{I})),
\end{equation}
where the global textual feature $\mathrm{M}_T$ is used as the query and the filtered visual feature $\hat{\mathrm{H}_{I}}$ serves as key and value.

Finally, on the basis of $\mathrm{H}_f$, we construct a classifier based on a two-layer multi-layer perception (MLP) to produce a keyphrase distribution $d_{cla}$ as follows:
\begin{equation}
    d_{cla}=\text{Softmax}(\text{MLP}(\text{H}_{f})).\label{eq:cl}
\end{equation}

\subsubsection{\textbf{Keyphrase Generation Module}} \

As implemented in \cite{DBLP:conf/emnlp/WangLLK20}, we introduce the pointer network \cite{DBLP:conf/acl/GuLLL16} to generate each keyphrase $y$ as a sequence. Typically, by equipping with an extended copy mechanism, this module models the token-level generation probability $p(y_j)$ at each timestep $j$ as the weighted sum of two types of probabilities:

\textbf{\emph{Prediction probability $p_{p}(y_j)$.}} To model this probability, we update the decoder hidden state $s_j$ as follows:
\begin{align}
\qquad s_j&=\mathrm{GRU}(y_{j-1},s_{j-1},c_j), \\
\qquad c_j&=\sum_{i=1}^{|X_T|}\alpha_{j,i}h_i, \\
\qquad \alpha_{j,i}&=\text{Softmax}(V_{\alpha}^{\top}\text{tanh}(W_{\alpha}[s_j;h_i])), 
\end{align}
where $y_{j-1}$ is the output at timestep $j\text{-}1$, $c_j$ is the context vector, $\alpha_{j,i}$ is the normalized weight that measures the compatibility between $s_j$ and $h_i$, $V_{\alpha}$ and $V_{\alpha}$ are learnable parameter matrices. 

Next, we further introduce the fusion vector $\text{H}_{f}$ to produce a token distribution $p_{p}(y_j)$ as follows:
\begin{equation}
    \qquad p_{p}(y_j)=\text{Softmax}(W_p[y_{j-1};s_j;c_j+\text{H}_{f}]),
\end{equation}
where $W_p$ is a learnable parameter matrix.

\textbf{\emph{Copy probability $P_{c}(y_j)$.}} To generate better keyphrases, we also adopt an extended copy mechanism to simultaneously leverage the words of concatenated input text $X_T$ and the classifier predictions $d_{cla}$.

Specifically, we first retrieve the top-5 classifier predictions and transform each prediction into a sequence of words $\mathbf{w}=w_1,…,w_{|\mathbf{w}|}$. Afterwards, we use a softmax function to normalize the corresponding classification logits into word-level distributions $\left\{\beta_k\right\}_{k=1}^{|\mathbf{w}|}$. Finally, we define the copy probability $p_{c}(y_j)$ as
\begin{equation}
    \qquad p_{c}(y_j)= \lambda_c \cdot\sum_{i:x_i=y_j}^{|X_T|}\alpha_{j,i} +(1-\lambda_c)\cdot\sum_{k:w_k=y_j}^{|\mathbf{w}|}\beta_k,
\end{equation}
where $\lambda_c$ is a hyper-parameter used to decide whether to copy from the concatenated input text or the classification predictions.

With the above two kinds of probabilities, we obtain the generation probability $p(y_j)$ as follows:
\begin{align}
    p(y_j)={\lambda p_{p}(y_j)+(1-\lambda})p_{c}(y_j), \\
    \lambda=\text{Sigmoid}(W_{\lambda}[y_{j-1};s_j;c_j+\text{H}_{f}]),
\end{align}
where $\lambda$ is a soft switch and $W_{\lambda}$ is a learnable parameter matrix. 

\subsection{Training Framework}
\begin{sloppypar}
We propose a two-stage training framework to train our model.

\textbf{Stage 1.} During this stage, we first pre-train the multi-modal feature encoding module, image noise filtering module and keyphrase classification module. To this end, we define the following training objective involving three loss items:
\begin{equation}
\mathcal{L}_{1}=\mathcal{L}_{itm}+\mathcal{L}_{irtm}+\mathcal{L}_{cla},
\end{equation}
where $\mathcal{L}_{itm}$, $\mathcal{L}_{irtm}$, $\mathcal{L}_{cla}$ are loss items proposed for three tasks. We will describe in detail these three losses, respectively.

\emph{The loss item for image-text matching: $\mathcal{L}_{itm}$.}
As described in previously, we introduce an image-text matching task to perform coarse-granularity image noise filtering. Given an additional dataset $D_{itm}=\left\{(X_T,X_I)\right\}$, we define the following cross-entropy loss: 
\begin{equation}
 \mathcal{L}_{itm}=-\sum_{(X_T, X_I)\in D_{itm}}\text{log}(p_{itm}(X_T,X_I)),
\end{equation}
where $p_{itm}(*)$ is the probability of correct classification.

\emph{The loss item for image region-text matching: $\mathcal{L}_{irtm}$.} As mentioned above, for each training text-image pair $(X_S, X_I, {y}) \in{D}$, we introduce a correlation matrix $\mathrm{A}$ to perform fine-granularity image noise filtering. To accurately model $\mathrm{A}$, we encode the ground-truth keyphrases and calculate the correlation score between each region of input image and ground-truth keyphrases according to Equation \ref{eq:matrix}, forming a correlation matrix $\mathrm{A}_{gt}$. Afterwards, we use $\mathrm{A}_{gt}$ as supervisory signals to train $\mathrm{A}$ by introducing a MSE(Mean Squared Error) loss to minimize their divergence:
\begin{equation}
 \mathcal{L}_{irtm}=\sum_{(X_S, X_I) \in{D}}\text{MSE}(\mathrm{A},\mathrm{A}_{gt}).
\end{equation}


\emph{The loss item for keyphrase classification: $\mathcal{L}_{cla}$.} To train the previously-mentioned keyphrase classifier, we define the following standard cross-entropy loss:
\begin{equation}
 \mathcal{L}_{cla}=-\sum_{(X_S, X_I, {y}) \in{D}}\text{log}(d_{cla}),
\end{equation}
where $d_{cla}$ denotes the predictions of keyphrase classification, defined as Equation \ref{eq:cl}.

\textbf{Stage 2.} In this stage, we optimize the model for the keyphrase generation task. Following common practice \cite{DBLP:conf/acl/MengZHHBC17}, we design $\mathcal{L}_{2}$ as a token-level cross-entropy loss:
\begin{equation}
 \mathcal{L}_{2}=-\sum_{(X_S, X_I, {y})\in{D}}\sum_{j=1}^{|y|}\text{log}(p(y_j)).
\end{equation}


\end{sloppypar}

\section{EXPERIMENT}
\subsection{Setup}
\subsubsection{Dataset} \

In our experiments, we use two datasets.  
One is the TRC dataset\footnote{https://github.com/danielpreotiuc/text-image-relationship/}, which is used to train the model via the image-text matching task.
The other is the dataset for multi-modal keyphrase generation collected by \citet{DBLP:conf/emnlp/WangLLK20}.
This dataset includes 53,701 English tweets, each of which comprises a distinct text-image pair, with user-annotated hashtags serving as keyphrases.
The detailed statistics of these datasets are given in Appendix A.


\subsubsection{Implementation Details} \

To ensure fair comparisons, in the experiments, we use the setting used in \cite{DBLP:conf/emnlp/WangLLK20} which is our most important baseline. Specifically, we select the top 45K most frequent words as the vocabulary for keyphrase generation and 4,262 keyphrases of the training data as candidate ones in the classifier. When constructing our encoder and decoder, we initialize the input word embeddings with 200-dimensional GloVe \cite{DBLP:conf/emnlp/PenningtonSM14} ones, and set their hidden state dimensions as 300. To encode the input image, we use 49 grid-level VGG features, where each grid is represented as a 512-dimensional vector.
During training, we use Adam \cite{DBLP:journals/corr/KingmaB14} to optimize the model, with an initial learning rate of $10^{-3}$. Additionally, we perform dropout\cite{DBLP:journals/jmlr/SrivastavaHKSS14} with a rate of 0.1 to enhance the robustness of our model. Particularly, we employ early stopping to stop the model training according to the performance on the validation dataset. During inference, we apply beam search with a beam size of 10 to produce a ranked list of keyphrases. We conduct the experiments repeat five times using different random seeds, and report the averaged results. The experimental results we report are obtained by repeating five times with different seeds and then averaging the values.

\subsubsection{Baseline} \

We compare our model with various baselines, which can be roughly classified into the following three categories:
\begin{itemize}
    \item{\textbf{Image-only models.}} In this category, we consider two models.  1) \textbf{VGG}. This model utilizes the pre-trained VGG encoder to obtain visual features, which are then fed into a classifier for keyphrase prediction. 2) \textbf{BUTD} \cite{DBLP:conf/cvpr/00010BT0GZ18}. It first uses a bottom-up attention to detect objects and then extract their visual features for keyphrase classification.
    \item{\textbf{Text-only models.}} This category of models can be further divided into classification-based (CLA) ones and generation-based (GEN) ones. The typical models in the former mainly include the following three models. 1) \textbf{AVG}. This model simply leverages the average-pooling representation of textual features for keyphrase classification. 2) \textbf{MAX}. It uses the max-pooling representation of textual features to perform keyphrase classification. 3) \textbf{TMN} \cite{DBLP:conf/emnlp/ZengLSGLK18}, which introduces a topic memory network encoding latent topic representations for keyphrase classification. 
    Besides, the dominant GEN models mainly include the following five models. 1) \textbf{ATT} \cite{DBLP:journals/corr/BahdanauCB14}. It is based on an attention-based Seq2Seq generation framework. 2) \textbf{COPY} \cite{DBLP:conf/acl/SeeLM17}, which introduces a pointer network for keyphrase generation. 3) \textbf{TOPIC} \cite{DBLP:conf/acl/WangLCKLS19}. It is a topic-aware model that allows end-to-end learning of latent topic modeling and keyphrase generation. 4) \textbf{ONE2SEQ} \cite{DBLP:conf/acl/YuanWMTBHT20}. This model outputs an ordered keyphrase given the input text, while our model outputs one keyphrase at each timestep. 5) \textbf{ONE2SET} \cite{DBLP:conf/acl/YeGL0Z20}. It models keyphrase generation as a generation task of a keyphrase set, where keyphrases are individually generated in parallel. Please note that we use Transformer to build ONE2SEQ and ONE2SET, according to the results reported in \cite{DBLP:conf/acl/YeGL0Z20}.\footnote{To build ONE2SEQ and ONE2SET, we use the source code released at https://github.com/jiacheng-ye/kg\_one2set.}
    \item{\textbf{Text-image models.}} The dominant models in this category involve 1) \textbf{CO-ATT} \cite{DBLP:conf/ijcai/ZhangWHHG17}, which designs a co-attention network to learn token-aware visual representations for multi-modal hashtag recommendation. 2) \textbf{BAN} \cite{DBLP:conf/nips/KimJZ18}, which uses a bilinear attention network to capture bilinear interactions among visual features and textual features for keyphrase generation. 3) \textbf{FLAVA} \cite{DBLP:conf/cvpr/SinghHGCGRK22}, a language and vision alignment model, effectively learns representations from both multimodal and unimodal data, making it widely utilized in various multimodal tasks.
    4) \textbf{$\text{M}^3\text{H-ATT}$} \cite{DBLP:conf/emnlp/WangLLK20}. This model achieves SOTA performance in multi-modal keyphrase generation. Typically, it proposes a multi-modality multi-head attention mechanism and jointly models keyphrase classification and generation above. Likewise, we report the performance of the variant \textbf{$\text{M}^3\text{H-ATT}$(TF)} that is based on Transformer and \textbf{$\text{M}^3\text{H-ATT}$(PO)} that adopts PaddleOCR to perform OCR, which is similar to our method.
\end{itemize}

Following previous studies \cite{DBLP:conf/acl/MengZHHBC17,DBLP:conf/emnlp/WangLLK20}, we use the commonly-used macro-average F1@K to evaluate the model performance, where $\text{K}$ is 1 or 3.
Besides, as implemented in \cite{DBLP:conf/naacl/ChenCLBK19}, we measure the keyphrase orders with the mean average precision (MAP) for the top-5 predictions.
\begin{table}
\caption{Performance comparison for multi-modal keyphrase generation task. $\ast$ indicates the results are directly cited from \cite{DBLP:conf/emnlp/WangLLK20}. $\dagger$ indicates significant at p<0.01 over $\mathbf{M^3H}$-ATT with 1,000 booststrap tests. Note that ONE2SEQ, ONE2SET, and $\mathbf{M^3H}$-ATT(TF) are based on Transformer.}
  \label{tab:1}
  \begin{tabular}{llll}
    \toprule
    Models&F1@1&F1@3&MAP@5\\
    \midrule
    \multicolumn{4}{c}{\textit{Image-only models}} \\
    \midrule
    VGG$^{\ast }$           & 15.69    & 13.67    & 19.70    \\
    BUTD$^{\ast }$ \cite{DBLP:conf/cvpr/00010BT0GZ18}          & 20.02    & 16.97    & 24.73    \\
    \midrule
    \multicolumn{4}{c}{\textit{Text-only models}}  \\
    \midrule
    AVG$^{\ast }$            & 35.96    & 27.59    & 41.84    \\
    MAX$^{\ast }$           & 38.33    & 28.84    & 44.15    \\
    TMN$^{\ast }$ \cite{DBLP:conf/emnlp/ZengLSGLK18}          & 40.33    & 30.07    & 46.28    \\
    ATT$^{\ast }$ \cite{DBLP:journals/corr/BahdanauCB14}          & 38.36    & 27.83    & 43.35    \\
    COPY$^{\ast }$ \cite{DBLP:conf/acl/SeeLM17}         & 42.10    & 29.91    & 46.94    \\
    TOPIC$^{\ast }$ \cite{DBLP:conf/acl/WangLCKLS19}        & 43.17    & 30.73    & 48.07    \\
    ONE2SEQ \cite{DBLP:conf/acl/YuanWMTBHT20}        & 38.05    & 28.41    & 43.10    \\
    ONE2SET \cite{DBLP:conf/acl/YeGL0Z20}        & 36.36    & 33.75    & 37.47    \\
    \midrule
    \multicolumn{4}{c}{\textit{Text-image models}} \\
    \midrule
    CO-ATT$^{\ast }$ \cite{DBLP:conf/ijcai/ZhangWHHG17}       & 42.12    & 31.55    & 48.39    \\
    BAN$^{\ast }$ \cite{DBLP:conf/nips/KimJZ18}          & 38.73    & 29.68    & 45.03    \\
    FLAVA \cite{DBLP:conf/cvpr/SinghHGCGRK22}	&46.05	&31.23	&49.30 \\
    
    $\mathrm{M^3H}$-ATT$^{\ast }$ \cite{DBLP:conf/emnlp/WangLLK20}& 47.06    & 33.11    & 52.07    \\
    $\mathrm{M^3H}$-ATT(TF) \cite{DBLP:conf/emnlp/WangLLK20}& 40.08    & 29.78    & 46.09    \\
    $\mathrm{M^3H}$-ATT(PO)\cite{DBLP:conf/emnlp/WangLLK20} 	& 47.13	& 32.66	& 51.56 \\
    \midrule
    \multicolumn{4}{c}{\textit{Our text-image model}} \\
    \midrule
    Our Model     & \textbf{48.19}$^{\dagger}$    & \textbf{33.86}$^{\dagger}$    & \textbf{53.28}$^{\dagger}$   \\
  \bottomrule
\end{tabular}
\vspace{-2mm}
\end{table}

\subsection{Main Results} \
Table \ref{tab:1} shows the performance of our model and baselines on the dataset collected by \citet{DBLP:conf/emnlp/WangLLK20}. Here we can obtain the following conclusions:

First, we report the performance of $\mathrm{M^3H}$-ATT and its variant $\mathrm{M^3H}$-ATT(TF), which are constructed based on GRU and Transformer, respectively, respectively. We can clearly find that $\mathrm{M^3H}$-ATT(TF) is significantly inferior to $\mathrm{M^3H}$-ATT. We believe this result is reasonable, also echoing the conclusion mentioned in previous studies \cite{DBLP:conf/acl/FosterVUMKFJSWB18, DBLP:journals/corr/abs-2207-02098} that RNN-style models may outperform Transform-style ones in the low-resource scenarios, especially formal language tasks.
Hence, in the subsequent experiments, we mainly focus on GRU-style models.

Second, our model surpasses all baselines in terms of all metrics. Specifically, our model outperforms $\mathrm{M^3H}$-ATT by 1.2 points in terms of F1@1, 0.8 points in terms of F1@3, and 1.1 points in terms of MAP@5. Note that $\mathrm{M^3H}$-ATT is the current SOTA model in multi-modal keyphrase generation. This result strongly confirms the effectiveness of our model.

Third, the multi-modal models outperform both image-only and text-only models, echoing the results reported in \cite{DBLP:conf/emnlp/WangLLK20}. Notably, our model exhibits superior performance compared to the ONE2SEQ and ONE2SET, they are recently developed models that perform well on text-only keyphrase generation. Thus, we confirm that the complementarity between image and text enables our model to effectively capture crucial information for multi-modal keyphrase generation.

Finally, the text-only models perform better than the image-only ones, showing that each input text provides more cues than input image, and therefore can contribute more to keyphrase generation. For this result, we speculate that the inferior performance of image-only models may be attributed to the feature sparsity and noise in the input image, making it challenging for models to acquire effective features.


\begin{table}
\caption{Experimental results of ablation study.}
  \label{tab:2}
   \begin{tabular}{llll}
    \toprule
    Models&F1@1&F1@3&MAP@5\\
    \midrule
     Our Model           & \textbf{48.19}    & \textbf{33.86}    & \textbf{53.28}   \\
    \hline
      \quad \textit{w/o $\mathcal{L}_{itm}$}         & 47.76    & 33.66    & 52.94    \\
      \quad \textit{w/o $\mathcal{L}_{irtm}$}         & 48.03    & 33.46    & 52.88    \\
      \quad \textit{w/o $\mathcal{L}_{cla}$}         & 46.51    & 33.33    & 52.15    \\ 
      \quad \textit{w/o visual entities}              & 47.42    & 33.10    & 52.20    \\
      \quad \textit{w/o ocr text}                     & 46.72    & 32.95    & 51.70    \\
      \quad \textit{w/o image-text matching}  & 47.60   & 33.35    & 52.69    \\
      \quad \textit{w/o image region-text matching}    & 47.58   & 33.26    &52.58    \\ 
      \quad \textit{w/o image noise filtering module}    & 47.68   & 33.12    &52.48    \\ 
  \bottomrule
\end{tabular}
\vspace{-3mm}
\end{table}

\begin{figure*}[h]
  \centering
  \includegraphics[width=\linewidth]{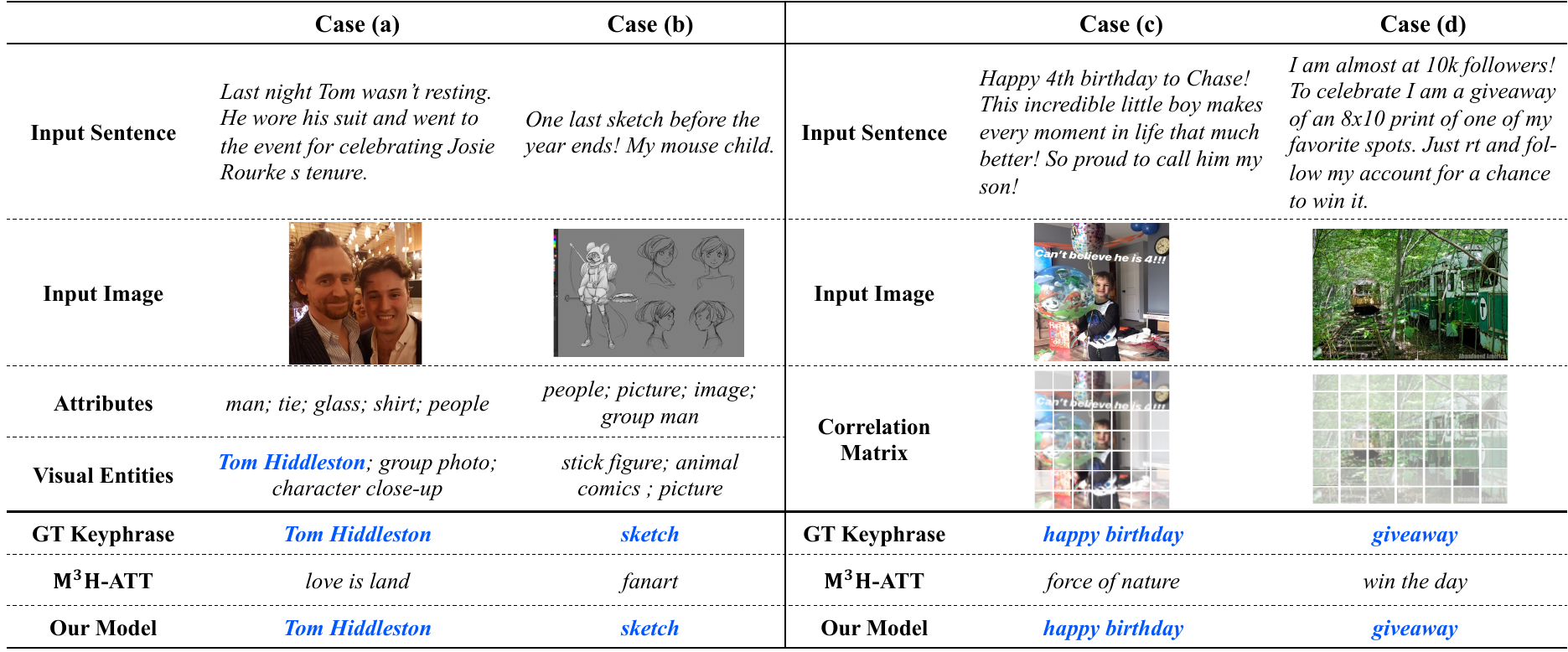}
  \caption{Examples of Multi-modal Keyphrase Generation. }
  \Description{}
  \label{fig:example1}
  \vspace{-4mm}
\end{figure*}
\subsection{Ablation Study} \
To investigate the effects of different factors on our model, we conduct an ablation study in Table \ref{tab:2}. Here, we consider the following variants:
\begin{itemize}
    \item{}{\emph{w/o $\mathcal{L}_{itm}$.}} Note that the loss item $\mathcal{L}_{itm}$ is proposed for the image-text matching strategy. In this way, this variant conducts coarse-granularity image noise filtering without supervisory information. The result in line 3 shows that it causes a performance decline.
    \item{}{\emph{w/o $\mathcal{L}_{irtm}$.}} In this variant, we remove the loss item $\mathcal{L}_{irtm}$ from the whole training objective. Thus, this variant does not use the correlation matrix $\text{A}_{gt}$ between ground-truth keyphrases and image regions to supervise the training of the correlation matrix A. As shown in line 4, the degradation of the model performance confirms our hypothesis that the correlation matrix $\text{A}_{gt}$ can guide the model to focus on key regions, and thus benefit the model performance.
    \item{}{\emph{w/o $\mathcal{L}_{cla}$.}} When removing this loss item, we discard the keyphrase classification task and thus directly adopt the conventional pointer network that only copies the words of the concatenated input text to generate keyphrases. According to the result shown in line 5, we can draw the conclusion that this classification task indeed significantly improves the model performance.
    \item{}\emph{w/o visual entities.} In this variant, we do not use the extracted visual entities. From line 6, we find that the model performance drops drastically. This result confirms our hypothesis that visual entities can provide effective supplementary information for keyphrase generation.
    \item{}{\emph{w/o ocr text.}} This variant does not use the OCR text during keyphrase predictions. The result presented in line 7 shows that the model exhibits performance degradation, suggesting that the OCR text is also useful for keyphrase generation.
    \item{}{\emph{w/o image-text matching.}} When constructing this variant, we discard the matching score $S_c$ to smooth $\mathrm{A}$ (See Equation \ref{eq:matrix}). Correspondingly, we remove the loss item $\mathcal{L}_{itm}$ from the whole training objective. As shown in line 8, this removal leads to a significant performance drop of 0.4 percentage points on the F1 score, demonstrating the crucial role of this module.
    \item{}{\emph{w/o image region-text matching.}} In this variant, we remove the text-image correlation matrix used for fine-granularity image noise filtering. Consequently, the loss item $\mathcal{L}_{irtm}$ is also removed from the whole training objective. From line 9, we can observe that filtering irrelevant image regions before multi-modal fusion is beneficial to improve model performance.
    \item{}{\emph{w/o image noise filtering module.}} We remove the entire image noise filtering module in this variant. Correspondingly, we remove the loss items $\mathcal{L}_{itm}$ and $\mathcal{L}_{irtm}$. According to the result shown in line 10, the drop in performance clearly justifies the necessity of our noise filtering module.

    
\end{itemize}

\subsection{Case Study} \
To further demonstrate the effectiveness of our model, we compare the keyphrases generated by our model and $\text{M}^3$H-ATT, which is our most important baseline. Some selected cases are shown in Figure \ref{fig:example1}.

As shown in case (a), we can use Baidu API to obtain a visual entity ``\emph{Tom Hiddleston}''. Obviously, compared with the image attributes, which indicate that there are two men in the input image, this visual entity can inform our model with the man’s name, which is more suitable to become a keyphrase.
Thus, our model is able to correctly produce the keyphrase ``\emph{Tom Hiddleston}''. On the contrary, it is difficult for $\mathrm{M^3H}$-ATT to successfully predict this keyphrase, since its OCR text and original input sentence do not mention the man's name. In addition to explicitly providing the possible keyphrases, visual entities can enable our model to exploit more related information about the input image, leading to better predictions. For example, in case (b), the extracted visual entities such as ``\emph{stick figure}", ``\emph{gongbi}", and ``\emph{comics}", are semantically related to the keyphrase ``\emph{sketch}''. However, these visual entities can only be utilized by our model, but not $\mathrm{M^3H}$-ATT.

We further analyze the correlation matrix $\mathrm{A}$ (See Equation \ref{eq:matrix}) to investigate the effect of multi-granularity image noise filtering on our model. Back to case (c), we observe that the keyphrase ``\emph{happy birthday}" is closely related to the input image. According to the correlation matrix $\mathrm{A}$, we can find that our model pays more attention to the key regions that are related to the input sentence, leading to the improvement of model performance. In contrast, in case (d), we find that multi-granularity image noise filtering enables our model to focus little on all image regions. In this way, our model relies more heavily on the input textual information for keyphrase generation. This result is reasonable, since the ground-truth keyphrase ``\emph{giveaway}'' is not related to the input image in this case.

\section{CONCLUSION}
In this paper, we have proposed a novel multi-modal keyphrase generation model that significantly extends the current SOTA model in two aspects.
First, we supplement the model input with external visual entities, 
which enhances the semantic alignment between image and text.
Second, we successively perform image-text matching and image region-text matching to effectively filter image noise.
We conduct several groups of experiments on the commonly-used dataset.
Experimental results and in-depth analyses verify the effectiveness of our model.

In the future, we will apply our model to other multi-modal tasks such as multi-modal machine translation \cite{ DBLP:conf/mm/LinMSYYGZL20, DBLP:conf/acl/YinMSZYZL20, DBLP:journals/isci/SuCJZLGWL21, DBLP:conf/acl/LanYLZ0WHS23, DBLP:conf/acl/KangHP0SCWHS23} and text summarization \cite{DBLP:conf/acl/Jiang00SZ23, DBLP:conf/acl/LiangMXW0023}, so as to further investigate its effectiveness and generality.

\begin{acks}
The project was supported by  
National Key Research and Development Program of China (No. 2020AAA0108004), 
National Natural Science Foundation of China (No. 62276219), 
and Natural Science Foundation of Fujian Province of China (No. 2020J06001 and No. 2021J01003). 
We also thank the reviewers for their insightful comments. 
\end{acks}

\balance
\bibliographystyle{ACM-Reference-Format}
\balance
\bibliography{sample-base}


\begin{thebibliography}{55}


\ifx \showCODEN    \undefined \def \showCODEN     #1{\unskip}     \fi
\ifx \showDOI      \undefined \def \showDOI       #1{#1}\fi
\ifx \showISBNx    \undefined \def \showISBNx     #1{\unskip}     \fi
\ifx \showISBNxiii \undefined \def \showISBNxiii  #1{\unskip}     \fi
\ifx \showISSN     \undefined \def \showISSN      #1{\unskip}     \fi
\ifx \showLCCN     \undefined \def \showLCCN      #1{\unskip}     \fi
\ifx \shownote     \undefined \def \shownote      #1{#1}          \fi
\ifx \showarticletitle \undefined \def \showarticletitle #1{#1}   \fi
\ifx \showURL      \undefined \def \showURL       {\relax}        \fi
\providecommand\bibfield[2]{#2}
\providecommand\bibinfo[2]{#2}
\providecommand\natexlab[1]{#1}
\providecommand\showeprint[2][]{arXiv:#2}

\bibitem[Anastasopoulos et~al\mbox{.}(2019)]%
        {DBLP:journals/corr/abs-1903-02930}
\bibfield{author}{\bibinfo{person}{Antonios Anastasopoulos}, \bibinfo{person}{Shankar Kumar}, {and} \bibinfo{person}{Hank Liao}.} \bibinfo{year}{2019}\natexlab{}.
\newblock \showarticletitle{Neural Language Modeling with Visual Features}.
\newblock \bibinfo{journal}{\emph{CoRR}}  \bibinfo{volume}{abs/1903.02930} (\bibinfo{year}{2019}).
\newblock
\showeprint[arXiv]{1903.02930}
\urldef\tempurl%
\url{http://arxiv.org/abs/1903.02930}
\showURL{%
\tempurl}


\bibitem[Anderson et~al\mbox{.}(2018)]%
        {DBLP:conf/cvpr/00010BT0GZ18}
\bibfield{author}{\bibinfo{person}{Peter Anderson}, \bibinfo{person}{Xiaodong He}, \bibinfo{person}{Chris Buehler}, \bibinfo{person}{Damien Teney}, \bibinfo{person}{Mark Johnson}, \bibinfo{person}{Stephen Gould}, {and} \bibinfo{person}{Lei Zhang}.} \bibinfo{year}{2018}\natexlab{}.
\newblock \showarticletitle{Bottom-Up and Top-Down Attention for Image Captioning and Visual Question Answering}. In \bibinfo{booktitle}{\emph{2018 {IEEE} Conference on Computer Vision and Pattern Recognition, {CVPR} 2018, Salt Lake City, UT, USA, June 18-22, 2018}}. \bibinfo{publisher}{Computer Vision Foundation / {IEEE} Computer Society}, \bibinfo{pages}{6077--6086}.
\newblock
\urldef\tempurl%
\url{https://doi.org/10.1109/CVPR.2018.00636}
\showDOI{\tempurl}


\bibitem[Bahdanau et~al\mbox{.}(2015)]%
        {DBLP:journals/corr/BahdanauCB14}
\bibfield{author}{\bibinfo{person}{Dzmitry Bahdanau}, \bibinfo{person}{Kyunghyun Cho}, {and} \bibinfo{person}{Yoshua Bengio}.} \bibinfo{year}{2015}\natexlab{}.
\newblock \showarticletitle{Neural Machine Translation by Jointly Learning to Align and Translate}. In \bibinfo{booktitle}{\emph{3rd International Conference on Learning Representations, {ICLR} 2015, San Diego, CA, USA, May 7-9, 2015, Conference Track Proceedings}}, \bibfield{editor}{\bibinfo{person}{Yoshua Bengio} {and} \bibinfo{person}{Yann LeCun}} (Eds.).
\newblock
\urldef\tempurl%
\url{http://arxiv.org/abs/1409.0473}
\showURL{%
\tempurl}


\bibitem[Ben{-}Younes et~al\mbox{.}(2019)]%
        {DBLP:conf/aaai/Ben-younesCTC19}
\bibfield{author}{\bibinfo{person}{H{\'{e}}di Ben{-}Younes}, \bibinfo{person}{R{\'{e}}mi Cad{\`{e}}ne}, \bibinfo{person}{Nicolas Thome}, {and} \bibinfo{person}{Matthieu Cord}.} \bibinfo{year}{2019}\natexlab{}.
\newblock \showarticletitle{{BLOCK:} Bilinear Superdiagonal Fusion for Visual Question Answering and Visual Relationship Detection}. In \bibinfo{booktitle}{\emph{The Thirty-Third {AAAI} Conference on Artificial Intelligence, {AAAI} 2019, Honolulu, Hawaii, USA, January 27 - February 1, 2019}}. \bibinfo{publisher}{{AAAI} Press}, \bibinfo{pages}{8102--8109}.
\newblock
\urldef\tempurl%
\url{https://doi.org/10.1609/aaai.v33i01.33018102}
\showDOI{\tempurl}


\bibitem[Campos et~al\mbox{.}(2018)]%
        {DBLP:conf/ecir/0001MPJNJ18a}
\bibfield{author}{\bibinfo{person}{Ricardo Campos}, \bibinfo{person}{V{\'{\i}}tor Mangaravite}, \bibinfo{person}{Arian Pasquali}, \bibinfo{person}{Al{\'{\i}}pio~M{\'{a}}rio Jorge}, \bibinfo{person}{C{\'{e}}lia Nunes}, {and} \bibinfo{person}{Adam Jatowt}.} \bibinfo{year}{2018}\natexlab{}.
\newblock \showarticletitle{YAKE! Collection-Independent Automatic Keyword Extractor}. In \bibinfo{booktitle}{\emph{Advances in Information Retrieval - 40th European Conference on {IR} Research, {ECIR} 2018, Grenoble, France, March 26-29, 2018, Proceedings}} \emph{(\bibinfo{series}{Lecture Notes in Computer Science}, Vol.~\bibinfo{volume}{10772})}. \bibinfo{publisher}{Springer}, \bibinfo{pages}{806--810}.
\newblock
\urldef\tempurl%
\url{https://doi.org/10.1007/978-3-319-76941-7\_80}
\showDOI{\tempurl}


\bibitem[Chen et~al\mbox{.}(2018b)]%
        {DBLP:conf/emnlp/ChenZ0YL18}
\bibfield{author}{\bibinfo{person}{Jun Chen}, \bibinfo{person}{Xiaoming Zhang}, \bibinfo{person}{Yu Wu}, \bibinfo{person}{Zhao Yan}, {and} \bibinfo{person}{Zhoujun Li}.} \bibinfo{year}{2018}\natexlab{b}.
\newblock \showarticletitle{Keyphrase Generation with Correlation Constraints}. In \bibinfo{booktitle}{\emph{Proceedings of the 2018 Conference on Empirical Methods in Natural Language Processing, Brussels, Belgium, October 31 - November 4, 2018}}, \bibfield{editor}{\bibinfo{person}{Ellen Riloff}, \bibinfo{person}{David Chiang}, \bibinfo{person}{Julia Hockenmaier}, {and} \bibinfo{person}{Jun'ichi Tsujii}} (Eds.). \bibinfo{publisher}{Association for Computational Linguistics}, \bibinfo{pages}{4057--4066}.
\newblock
\urldef\tempurl%
\url{https://doi.org/10.18653/v1/d18-1439}
\showDOI{\tempurl}


\bibitem[Chen et~al\mbox{.}(2018a)]%
        {DBLP:conf/acl/FosterVUMKFJSWB18}
\bibfield{author}{\bibinfo{person}{Mia~Xu Chen}, \bibinfo{person}{Orhan Firat}, \bibinfo{person}{Ankur Bapna}, \bibinfo{person}{Melvin Johnson}, \bibinfo{person}{Wolfgang Macherey}, \bibinfo{person}{George~F. Foster}, \bibinfo{person}{Llion Jones}, \bibinfo{person}{Mike Schuster}, \bibinfo{person}{Noam Shazeer}, \bibinfo{person}{Niki Parmar}, \bibinfo{person}{Ashish Vaswani}, \bibinfo{person}{Jakob Uszkoreit}, \bibinfo{person}{Lukasz Kaiser}, \bibinfo{person}{Zhifeng Chen}, \bibinfo{person}{Yonghui Wu}, {and} \bibinfo{person}{Macduff Hughes}.} \bibinfo{year}{2018}\natexlab{a}.
\newblock \showarticletitle{The Best of Both Worlds: Combining Recent Advances in Neural Machine Translation}. In \bibinfo{booktitle}{\emph{Proceedings of the 56th Annual Meeting of the Association for Computational Linguistics, {ACL} 2018, Melbourne, Australia, July 15-20, 2018, Volume 1: Long Papers}}. \bibinfo{publisher}{Association for Computational Linguistics}, \bibinfo{pages}{76--86}.
\newblock
\urldef\tempurl%
\url{https://doi.org/10.18653/v1/P18-1008}
\showDOI{\tempurl}


\bibitem[Chen et~al\mbox{.}(2019a)]%
        {DBLP:conf/naacl/ChenCLBK19}
\bibfield{author}{\bibinfo{person}{Wang Chen}, \bibinfo{person}{Hou~Pong Chan}, \bibinfo{person}{Piji Li}, \bibinfo{person}{Lidong Bing}, {and} \bibinfo{person}{Irwin King}.} \bibinfo{year}{2019}\natexlab{a}.
\newblock \showarticletitle{An Integrated Approach for Keyphrase Generation via Exploring the Power of Retrieval and Extraction}. In \bibinfo{booktitle}{\emph{Proceedings of the 2019 Conference of the North American Chapter of the Association for Computational Linguistics: Human Language Technologies, {NAACL-HLT} 2019, Minneapolis, MN, USA, June 2-7, 2019, Volume 1 (Long and Short Papers)}}. \bibinfo{publisher}{Association for Computational Linguistics}, \bibinfo{pages}{2846--2856}.
\newblock
\urldef\tempurl%
\url{https://doi.org/10.18653/v1/n19-1292}
\showDOI{\tempurl}


\bibitem[Chen et~al\mbox{.}(2020)]%
        {DBLP:conf/acl/ChenCLK20}
\bibfield{author}{\bibinfo{person}{Wang Chen}, \bibinfo{person}{Hou~Pong Chan}, \bibinfo{person}{Piji Li}, {and} \bibinfo{person}{Irwin King}.} \bibinfo{year}{2020}\natexlab{}.
\newblock \showarticletitle{Exclusive Hierarchical Decoding for Deep Keyphrase Generation}. In \bibinfo{booktitle}{\emph{Proceedings of the 58th Annual Meeting of the Association for Computational Linguistics, {ACL} 2020, Online, July 5-10, 2020}}. \bibinfo{publisher}{Association for Computational Linguistics}, \bibinfo{pages}{1095--1105}.
\newblock
\urldef\tempurl%
\url{https://doi.org/10.18653/v1/2020.acl-main.103}
\showDOI{\tempurl}


\bibitem[Chen et~al\mbox{.}(2019b)]%
        {DBLP:conf/aaai/ChenGZKL19}
\bibfield{author}{\bibinfo{person}{Wang Chen}, \bibinfo{person}{Yifan Gao}, \bibinfo{person}{Jiani Zhang}, \bibinfo{person}{Irwin King}, {and} \bibinfo{person}{Michael~R. Lyu}.} \bibinfo{year}{2019}\natexlab{b}.
\newblock \showarticletitle{Title-Guided Encoding for Keyphrase Generation}. In \bibinfo{booktitle}{\emph{The Thirty-Third {AAAI} Conference on Artificial Intelligence, {AAAI} 2019, The Thirty-First Innovative Applications of Artificial Intelligence Conference, {IAAI} 2019, The Ninth {AAAI} Symposium on Educational Advances in Artificial Intelligence, {EAAI} 2019, Honolulu, Hawaii, USA, January 27 - February 1, 2019}}. \bibinfo{publisher}{{AAAI} Press}, \bibinfo{pages}{6268--6275}.
\newblock
\urldef\tempurl%
\url{https://doi.org/10.1609/aaai.v33i01.33016268}
\showDOI{\tempurl}


\bibitem[Chen et~al\mbox{.}(2022)]%
        {DBLP:conf/naacl/ChenZLYDTHSC22}
\bibfield{author}{\bibinfo{person}{Xiang Chen}, \bibinfo{person}{Ningyu Zhang}, \bibinfo{person}{Lei Li}, \bibinfo{person}{Yunzhi Yao}, \bibinfo{person}{Shumin Deng}, \bibinfo{person}{Chuanqi Tan}, \bibinfo{person}{Fei Huang}, \bibinfo{person}{Luo Si}, {and} \bibinfo{person}{Huajun Chen}.} \bibinfo{year}{2022}\natexlab{}.
\newblock \showarticletitle{Good Visual Guidance Make {A} Better Extractor: Hierarchical Visual Prefix for Multimodal Entity and Relation Extraction}. In \bibinfo{booktitle}{\emph{Findings of the Association for Computational Linguistics: {NAACL} 2022, Seattle, WA, United States, July 10-15, 2022}}, \bibfield{editor}{\bibinfo{person}{Marine Carpuat}, \bibinfo{person}{Marie{-}Catherine de~Marneffe}, {and} \bibinfo{person}{Iv{\'{a}}n Vladimir~Meza Ru{\'{\i}}z}} (Eds.). \bibinfo{publisher}{Association for Computational Linguistics}, \bibinfo{pages}{1607--1618}.
\newblock
\urldef\tempurl%
\url{https://doi.org/10.18653/v1/2022.findings-naacl.121}
\showDOI{\tempurl}


\bibitem[Del{\'{e}}tang et~al\mbox{.}(2022)]%
        {DBLP:journals/corr/abs-2207-02098}
\bibfield{author}{\bibinfo{person}{Gr{\'{e}}goire Del{\'{e}}tang}, \bibinfo{person}{Anian Ruoss}, \bibinfo{person}{Jordi Grau{-}Moya}, \bibinfo{person}{Tim Genewein}, \bibinfo{person}{Li~Kevin Wenliang}, \bibinfo{person}{Elliot Catt}, \bibinfo{person}{Marcus Hutter}, \bibinfo{person}{Shane Legg}, {and} \bibinfo{person}{Pedro~A. Ortega}.} \bibinfo{year}{2022}\natexlab{}.
\newblock \showarticletitle{Neural Networks and the Chomsky Hierarchy}.
\newblock \bibinfo{journal}{\emph{CoRR}}  \bibinfo{volume}{abs/2207.02098} (\bibinfo{year}{2022}).
\newblock
\urldef\tempurl%
\url{https://doi.org/10.48550/arXiv.2207.02098}
\showDOI{\tempurl}
\showeprint[arXiv]{2207.02098}


\bibitem[El{-}Beltagy and Rafea(2009)]%
        {DBLP:journals/is/El-BeltagyR09}
\bibfield{author}{\bibinfo{person}{Samhaa~R. El{-}Beltagy} {and} \bibinfo{person}{Ahmed~A. Rafea}.} \bibinfo{year}{2009}\natexlab{}.
\newblock \showarticletitle{KP-Miner: {A} keyphrase extraction system for English and Arabic documents}.
\newblock \bibinfo{journal}{\emph{Inf. Syst.}} \bibinfo{volume}{34}, \bibinfo{number}{1} (\bibinfo{year}{2009}), \bibinfo{pages}{132--144}.
\newblock
\urldef\tempurl%
\url{https://doi.org/10.1016/j.is.2008.05.002}
\showDOI{\tempurl}


\bibitem[Gong and Zhang(2016)]%
        {DBLP:conf/ijcai/GongZ16}
\bibfield{author}{\bibinfo{person}{Yuyun Gong} {and} \bibinfo{person}{Qi Zhang}.} \bibinfo{year}{2016}\natexlab{}.
\newblock \showarticletitle{Hashtag Recommendation Using Attention-Based Convolutional Neural Network}. In \bibinfo{booktitle}{\emph{Proceedings of the Twenty-Fifth International Joint Conference on Artificial Intelligence, {IJCAI} 2016, New York, NY, USA, 9-15 July 2016}}, \bibfield{editor}{\bibinfo{person}{Subbarao Kambhampati}} (Ed.). \bibinfo{publisher}{{IJCAI/AAAI} Press}, \bibinfo{pages}{2782--2788}.
\newblock
\urldef\tempurl%
\url{http://www.ijcai.org/Abstract/16/395}
\showURL{%
\tempurl}


\bibitem[Gu et~al\mbox{.}(2016)]%
        {DBLP:conf/acl/GuLLL16}
\bibfield{author}{\bibinfo{person}{Jiatao Gu}, \bibinfo{person}{Zhengdong Lu}, \bibinfo{person}{Hang Li}, {and} \bibinfo{person}{Victor O.~K. Li}.} \bibinfo{year}{2016}\natexlab{}.
\newblock \showarticletitle{Incorporating Copying Mechanism in Sequence-to-Sequence Learning}. In \bibinfo{booktitle}{\emph{Proceedings of the 54th Annual Meeting of the Association for Computational Linguistics, {ACL} 2016, August 7-12, 2016, Berlin, Germany, Volume 1: Long Papers}}. \bibinfo{publisher}{The Association for Computer Linguistics}.
\newblock
\urldef\tempurl%
\url{https://doi.org/10.18653/v1/p16-1154}
\showDOI{\tempurl}


\bibitem[Jiang et~al\mbox{.}(2023)]%
        {DBLP:conf/acl/Jiang00SZ23}
\bibfield{author}{\bibinfo{person}{Chaoya Jiang}, \bibinfo{person}{Rui Xie}, \bibinfo{person}{Wei Ye}, \bibinfo{person}{Jinan Sun}, {and} \bibinfo{person}{Shikun Zhang}.} \bibinfo{year}{2023}\natexlab{}.
\newblock \showarticletitle{Exploiting Pseudo Image Captions for Multimodal Summarization}. In \bibinfo{booktitle}{\emph{Findings of the Association for Computational Linguistics: {ACL} 2023, Toronto, Canada, July 9-14, 2023}}, \bibfield{editor}{\bibinfo{person}{Anna Rogers}, \bibinfo{person}{Jordan~L. Boyd{-}Graber}, {and} \bibinfo{person}{Naoaki Okazaki}} (Eds.). \bibinfo{publisher}{Association for Computational Linguistics}, \bibinfo{pages}{161--175}.
\newblock
\urldef\tempurl%
\url{https://aclanthology.org/2023.findings-acl.12}
\showURL{%
\tempurl}


\bibitem[Joshi et~al\mbox{.}(2022)]%
        {DBLP:conf/naacl/JoshiBJSM22}
\bibfield{author}{\bibinfo{person}{Abhinav Joshi}, \bibinfo{person}{Ashwani Bhat}, \bibinfo{person}{Ayush Jain}, \bibinfo{person}{Atin~Vikram Singh}, {and} \bibinfo{person}{Ashutosh Modi}.} \bibinfo{year}{2022}\natexlab{}.
\newblock \showarticletitle{{COGMEN:} COntextualized {GNN} based Multimodal Emotion recognitioN}. In \bibinfo{booktitle}{\emph{Proceedings of the 2022 Conference of the North American Chapter of the Association for Computational Linguistics: Human Language Technologies, {NAACL} 2022, Seattle, WA, United States, July 10-15, 2022}}. \bibinfo{publisher}{Association for Computational Linguistics}, \bibinfo{pages}{4148--4164}.
\newblock
\urldef\tempurl%
\url{https://doi.org/10.18653/v1/2022.naacl-main.306}
\showDOI{\tempurl}


\bibitem[Kang et~al\mbox{.}(2023)]%
        {DBLP:conf/acl/KangHP0SCWHS23}
\bibfield{author}{\bibinfo{person}{Liyan Kang}, \bibinfo{person}{Luyang Huang}, \bibinfo{person}{Ningxin Peng}, \bibinfo{person}{Peihao Zhu}, \bibinfo{person}{Zewei Sun}, \bibinfo{person}{Shanbo Cheng}, \bibinfo{person}{Mingxuan Wang}, \bibinfo{person}{Degen Huang}, {and} \bibinfo{person}{Jinsong Su}.} \bibinfo{year}{2023}\natexlab{}.
\newblock \showarticletitle{BigVideo: {A} Large-scale Video Subtitle Translation Dataset for Multimodal Machine Translation}. In \bibinfo{booktitle}{\emph{Findings of the Association for Computational Linguistics: {ACL} 2023, Toronto, Canada, July 9-14, 2023}}, \bibfield{editor}{\bibinfo{person}{Anna Rogers}, \bibinfo{person}{Jordan~L. Boyd{-}Graber}, {and} \bibinfo{person}{Naoaki Okazaki}} (Eds.). \bibinfo{publisher}{Association for Computational Linguistics}, \bibinfo{pages}{8456--8473}.
\newblock
\urldef\tempurl%
\url{https://aclanthology.org/2023.findings-acl.535}
\showURL{%
\tempurl}


\bibitem[Kim et~al\mbox{.}(2018)]%
        {DBLP:conf/nips/KimJZ18}
\bibfield{author}{\bibinfo{person}{Jin{-}Hwa Kim}, \bibinfo{person}{Jaehyun Jun}, {and} \bibinfo{person}{Byoung{-}Tak Zhang}.} \bibinfo{year}{2018}\natexlab{}.
\newblock \showarticletitle{Bilinear Attention Networks}. In \bibinfo{booktitle}{\emph{Advances in Neural Information Processing Systems 31: Annual Conference on Neural Information Processing Systems 2018, NeurIPS 2018, December 3-8, 2018, Montr{\'{e}}al, Canada}}. \bibinfo{pages}{1571--1581}.
\newblock
\urldef\tempurl%
\url{https://proceedings.neurips.cc/paper/2018/hash/96ea64f3a1aa2fd00c72faacf0cb8ac9-Abstract.html}
\showURL{%
\tempurl}


\bibitem[Kim et~al\mbox{.}(2021)]%
        {DBLP:conf/icml/KimSK21}
\bibfield{author}{\bibinfo{person}{Wonjae Kim}, \bibinfo{person}{Bokyung Son}, {and} \bibinfo{person}{Ildoo Kim}.} \bibinfo{year}{2021}\natexlab{}.
\newblock \showarticletitle{ViLT: Vision-and-Language Transformer Without Convolution or Region Supervision}. In \bibinfo{booktitle}{\emph{Proceedings of the 38th International Conference on Machine Learning, {ICML} 2021, 18-24 July 2021, Virtual Event}} \emph{(\bibinfo{series}{Proceedings of Machine Learning Research}, Vol.~\bibinfo{volume}{139})}, \bibfield{editor}{\bibinfo{person}{Marina Meila} {and} \bibinfo{person}{Tong Zhang}} (Eds.). \bibinfo{publisher}{{PMLR}}, \bibinfo{pages}{5583--5594}.
\newblock
\urldef\tempurl%
\url{http://proceedings.mlr.press/v139/kim21k.html}
\showURL{%
\tempurl}


\bibitem[Kingma and Ba(2015)]%
        {DBLP:journals/corr/KingmaB14}
\bibfield{author}{\bibinfo{person}{Diederik~P. Kingma} {and} \bibinfo{person}{Jimmy Ba}.} \bibinfo{year}{2015}\natexlab{}.
\newblock \showarticletitle{Adam: {A} Method for Stochastic Optimization}. In \bibinfo{booktitle}{\emph{3rd International Conference on Learning Representations, {ICLR} 2015, San Diego, CA, USA, May 7-9, 2015, Conference Track Proceedings}}, \bibfield{editor}{\bibinfo{person}{Yoshua Bengio} {and} \bibinfo{person}{Yann LeCun}} (Eds.).
\newblock
\urldef\tempurl%
\url{http://arxiv.org/abs/1412.6980}
\showURL{%
\tempurl}


\bibitem[Lan et~al\mbox{.}(2023)]%
        {DBLP:conf/acl/LanYLZ0WHS23}
\bibfield{author}{\bibinfo{person}{Zhibin Lan}, \bibinfo{person}{Jiawei Yu}, \bibinfo{person}{Xiang Li}, \bibinfo{person}{Wen Zhang}, \bibinfo{person}{Jian Luan}, \bibinfo{person}{Bin Wang}, \bibinfo{person}{Degen Huang}, {and} \bibinfo{person}{Jinsong Su}.} \bibinfo{year}{2023}\natexlab{}.
\newblock \showarticletitle{Exploring Better Text Image Translation with Multimodal Codebook}. In \bibinfo{booktitle}{\emph{Proceedings of the 61st Annual Meeting of the Association for Computational Linguistics (Volume 1: Long Papers), {ACL} 2023, Toronto, Canada, July 9-14, 2023}}, \bibfield{editor}{\bibinfo{person}{Anna Rogers}, \bibinfo{person}{Jordan~L. Boyd{-}Graber}, {and} \bibinfo{person}{Naoaki Okazaki}} (Eds.). \bibinfo{publisher}{Association for Computational Linguistics}, \bibinfo{pages}{3479--3491}.
\newblock
\urldef\tempurl%
\url{https://aclanthology.org/2023.acl-long.192}
\showURL{%
\tempurl}


\bibitem[Liang et~al\mbox{.}(2023)]%
        {DBLP:conf/acl/LiangMXW0023}
\bibfield{author}{\bibinfo{person}{Yunlong Liang}, \bibinfo{person}{Fandong Meng}, \bibinfo{person}{Jinan Xu}, \bibinfo{person}{Jiaan Wang}, \bibinfo{person}{Yufeng Chen}, {and} \bibinfo{person}{Jie Zhou}.} \bibinfo{year}{2023}\natexlab{}.
\newblock \showarticletitle{Summary-Oriented Vision Modeling for Multimodal Abstractive Summarization}. In \bibinfo{booktitle}{\emph{Proceedings of the 61st Annual Meeting of the Association for Computational Linguistics (Volume 1: Long Papers), {ACL} 2023, Toronto, Canada, July 9-14, 2023}}, \bibfield{editor}{\bibinfo{person}{Anna Rogers}, \bibinfo{person}{Jordan~L. Boyd{-}Graber}, {and} \bibinfo{person}{Naoaki Okazaki}} (Eds.). \bibinfo{publisher}{Association for Computational Linguistics}, \bibinfo{pages}{2934--2951}.
\newblock
\urldef\tempurl%
\url{https://aclanthology.org/2023.acl-long.165}
\showURL{%
\tempurl}


\bibitem[Lin et~al\mbox{.}(2020)]%
        {DBLP:conf/mm/LinMSYYGZL20}
\bibfield{author}{\bibinfo{person}{Huan Lin}, \bibinfo{person}{Fandong Meng}, \bibinfo{person}{Jinsong Su}, \bibinfo{person}{Yongjing Yin}, \bibinfo{person}{Zhengyuan Yang}, \bibinfo{person}{Yubin Ge}, \bibinfo{person}{Jie Zhou}, {and} \bibinfo{person}{Jiebo Luo}.} \bibinfo{year}{2020}\natexlab{}.
\newblock \showarticletitle{Dynamic Context-guided Capsule Network for Multimodal Machine Translation}. In \bibinfo{booktitle}{\emph{{MM} '20: The 28th {ACM} International Conference on Multimedia, Virtual Event / Seattle, WA, USA, October 12-16, 2020}}, \bibfield{editor}{\bibinfo{person}{Chang~Wen Chen}, \bibinfo{person}{Rita Cucchiara}, \bibinfo{person}{Xian{-}Sheng Hua}, \bibinfo{person}{Guo{-}Jun Qi}, \bibinfo{person}{Elisa Ricci}, \bibinfo{person}{Zhengyou Zhang}, {and} \bibinfo{person}{Roger Zimmermann}} (Eds.). \bibinfo{publisher}{{ACM}}, \bibinfo{pages}{1320--1329}.
\newblock
\urldef\tempurl%
\url{https://doi.org/10.1145/3394171.3413715}
\showDOI{\tempurl}


\bibitem[Liu et~al\mbox{.}(2017)]%
        {DBLP:journals/corr/abs-1712-00559}
\bibfield{author}{\bibinfo{person}{Chenxi Liu}, \bibinfo{person}{Barret Zoph}, \bibinfo{person}{Jonathon Shlens}, \bibinfo{person}{Wei Hua}, \bibinfo{person}{Li{-}Jia Li}, \bibinfo{person}{Li Fei{-}Fei}, \bibinfo{person}{Alan~L. Yuille}, \bibinfo{person}{Jonathan Huang}, {and} \bibinfo{person}{Kevin Murphy}.} \bibinfo{year}{2017}\natexlab{}.
\newblock \showarticletitle{Progressive Neural Architecture Search}.
\newblock \bibinfo{journal}{\emph{CoRR}}  \bibinfo{volume}{abs/1712.00559} (\bibinfo{year}{2017}).
\newblock
\showeprint[arXiv]{1712.00559}
\urldef\tempurl%
\url{http://arxiv.org/abs/1712.00559}
\showURL{%
\tempurl}


\bibitem[Meng et~al\mbox{.}(2017)]%
        {DBLP:conf/acl/MengZHHBC17}
\bibfield{author}{\bibinfo{person}{Rui Meng}, \bibinfo{person}{Sanqiang Zhao}, \bibinfo{person}{Shuguang Han}, \bibinfo{person}{Daqing He}, \bibinfo{person}{Peter Brusilovsky}, {and} \bibinfo{person}{Yu Chi}.} \bibinfo{year}{2017}\natexlab{}.
\newblock \showarticletitle{Deep Keyphrase Generation}. In \bibinfo{booktitle}{\emph{Proceedings of the 55th Annual Meeting of the Association for Computational Linguistics, {ACL} 2017, Vancouver, Canada, July 30 - August 4, Volume 1: Long Papers}}. \bibinfo{publisher}{Association for Computational Linguistics}, \bibinfo{pages}{582--592}.
\newblock
\urldef\tempurl%
\url{https://doi.org/10.18653/v1/P17-1054}
\showDOI{\tempurl}


\bibitem[Nam et~al\mbox{.}(2017)]%
        {DBLP:conf/cvpr/NamHK17}
\bibfield{author}{\bibinfo{person}{Hyeonseob Nam}, \bibinfo{person}{Jung{-}Woo Ha}, {and} \bibinfo{person}{Jeonghee Kim}.} \bibinfo{year}{2017}\natexlab{}.
\newblock \showarticletitle{Dual Attention Networks for Multimodal Reasoning and Matching}. In \bibinfo{booktitle}{\emph{2017 {IEEE} Conference on Computer Vision and Pattern Recognition, {CVPR} 2017, Honolulu, HI, USA, July 21-26, 2017}}. \bibinfo{publisher}{{IEEE} Computer Society}, \bibinfo{pages}{2156--2164}.
\newblock
\urldef\tempurl%
\url{https://doi.org/10.1109/CVPR.2017.232}
\showDOI{\tempurl}


\bibitem[Noh et~al\mbox{.}(2016)]%
        {DBLP:conf/cvpr/NohSH16}
\bibfield{author}{\bibinfo{person}{Hyeonwoo Noh}, \bibinfo{person}{Paul~Hongsuck Seo}, {and} \bibinfo{person}{Bohyung Han}.} \bibinfo{year}{2016}\natexlab{}.
\newblock \showarticletitle{Image Question Answering Using Convolutional Neural Network with Dynamic Parameter Prediction}. In \bibinfo{booktitle}{\emph{2016 {IEEE} Conference on Computer Vision and Pattern Recognition, {CVPR} 2016, Las Vegas, NV, USA, June 27-30, 2016}}. \bibinfo{publisher}{{IEEE} Computer Society}, \bibinfo{pages}{30--38}.
\newblock
\urldef\tempurl%
\url{https://doi.org/10.1109/CVPR.2016.11}
\showDOI{\tempurl}


\bibitem[Pennington et~al\mbox{.}(2014)]%
        {DBLP:conf/emnlp/PenningtonSM14}
\bibfield{author}{\bibinfo{person}{Jeffrey Pennington}, \bibinfo{person}{Richard Socher}, {and} \bibinfo{person}{Christopher~D. Manning}.} \bibinfo{year}{2014}\natexlab{}.
\newblock \showarticletitle{Glove: Global Vectors for Word Representation}. In \bibinfo{booktitle}{\emph{Proceedings of the 2014 Conference on Empirical Methods in Natural Language Processing, {EMNLP} 2014, October 25-29, 2014, Doha, Qatar, {A} meeting of SIGDAT, a Special Interest Group of the {ACL}}}, \bibfield{editor}{\bibinfo{person}{Alessandro Moschitti}, \bibinfo{person}{Bo~Pang}, {and} \bibinfo{person}{Walter Daelemans}} (Eds.). \bibinfo{publisher}{{ACL}}, \bibinfo{pages}{1532--1543}.
\newblock
\urldef\tempurl%
\url{https://doi.org/10.3115/v1/d14-1162}
\showDOI{\tempurl}


\bibitem[P{\'{e}}rez{-}R{\'{u}}a et~al\mbox{.}(2018)]%
        {DBLP:conf/bmvc/Perez-RuaBP18}
\bibfield{author}{\bibinfo{person}{Juan{-}Manuel P{\'{e}}rez{-}R{\'{u}}a}, \bibinfo{person}{Moez Baccouche}, {and} \bibinfo{person}{St{\'{e}}phane Pateux}.} \bibinfo{year}{2018}\natexlab{}.
\newblock \showarticletitle{Efficient Progressive Neural Architecture Search}. In \bibinfo{booktitle}{\emph{British Machine Vision Conference 2018, {BMVC} 2018, Newcastle, UK, September 3-6, 2018}}. \bibinfo{publisher}{{BMVA} Press}, \bibinfo{pages}{150}.
\newblock
\urldef\tempurl%
\url{http://bmvc2018.org/contents/papers/0291.pdf}
\showURL{%
\tempurl}


\bibitem[P{\'{e}}rez{-}R{\'{u}}a et~al\mbox{.}(2019)]%
        {DBLP:conf/cvpr/Perez-RuaVPBJ19}
\bibfield{author}{\bibinfo{person}{Juan{-}Manuel P{\'{e}}rez{-}R{\'{u}}a}, \bibinfo{person}{Valentin Vielzeuf}, \bibinfo{person}{St{\'{e}}phane Pateux}, \bibinfo{person}{Moez Baccouche}, {and} \bibinfo{person}{Fr{\'{e}}d{\'{e}}ric Jurie}.} \bibinfo{year}{2019}\natexlab{}.
\newblock \showarticletitle{{MFAS:} Multimodal Fusion Architecture Search}. In \bibinfo{booktitle}{\emph{{IEEE} Conference on Computer Vision and Pattern Recognition, {CVPR} 2019, Long Beach, CA, USA, June 16-20, 2019}}. \bibinfo{publisher}{Computer Vision Foundation / {IEEE}}, \bibinfo{pages}{6966--6975}.
\newblock
\urldef\tempurl%
\url{https://doi.org/10.1109/CVPR.2019.00713}
\showDOI{\tempurl}


\bibitem[Salton and Buckley(1988)]%
        {DBLP:journals/ipm/SaltonB88}
\bibfield{author}{\bibinfo{person}{Gerard Salton} {and} \bibinfo{person}{Chris Buckley}.} \bibinfo{year}{1988}\natexlab{}.
\newblock \showarticletitle{Term-Weighting Approaches in Automatic Text Retrieval}.
\newblock \bibinfo{journal}{\emph{Inf. Process. Manag.}} \bibinfo{volume}{24}, \bibinfo{number}{5} (\bibinfo{year}{1988}), \bibinfo{pages}{513--523}.
\newblock
\urldef\tempurl%
\url{https://doi.org/10.1016/0306-4573(88)90021-0}
\showDOI{\tempurl}


\bibitem[Sedhai and Sun(2014)]%
        {DBLP:conf/sigir/SedhaiS14}
\bibfield{author}{\bibinfo{person}{Surendra Sedhai} {and} \bibinfo{person}{Aixin Sun}.} \bibinfo{year}{2014}\natexlab{}.
\newblock \showarticletitle{Hashtag recommendation for hyperlinked tweets}. In \bibinfo{booktitle}{\emph{The 37th International {ACM} {SIGIR} Conference on Research and Development in Information Retrieval, {SIGIR} '14, Gold Coast , QLD, Australia - July 06 - 11, 2014}}, \bibfield{editor}{\bibinfo{person}{Shlomo Geva}, \bibinfo{person}{Andrew Trotman}, \bibinfo{person}{Peter Bruza}, \bibinfo{person}{Charles L.~A. Clarke}, {and} \bibinfo{person}{Kalervo J{\"{a}}rvelin}} (Eds.). \bibinfo{publisher}{{ACM}}, \bibinfo{pages}{831--834}.
\newblock
\urldef\tempurl%
\url{https://doi.org/10.1145/2600428.2609452}
\showDOI{\tempurl}


\bibitem[See et~al\mbox{.}(2017)]%
        {DBLP:conf/acl/SeeLM17}
\bibfield{author}{\bibinfo{person}{Abigail See}, \bibinfo{person}{Peter~J. Liu}, {and} \bibinfo{person}{Christopher~D. Manning}.} \bibinfo{year}{2017}\natexlab{}.
\newblock \showarticletitle{Get To The Point: Summarization with Pointer-Generator Networks}. In \bibinfo{booktitle}{\emph{Proceedings of the 55th Annual Meeting of the Association for Computational Linguistics, {ACL} 2017, Vancouver, Canada, July 30 - August 4, Volume 1: Long Papers}}. \bibinfo{publisher}{Association for Computational Linguistics}, \bibinfo{pages}{1073--1083}.
\newblock
\urldef\tempurl%
\url{https://doi.org/10.18653/v1/P17-1099}
\showDOI{\tempurl}


\bibitem[Simonyan and Zisserman(2015)]%
        {DBLP:journals/corr/SimonyanZ14a}
\bibfield{author}{\bibinfo{person}{Karen Simonyan} {and} \bibinfo{person}{Andrew Zisserman}.} \bibinfo{year}{2015}\natexlab{}.
\newblock \showarticletitle{Very Deep Convolutional Networks for Large-Scale Image Recognition}. In \bibinfo{booktitle}{\emph{3rd International Conference on Learning Representations, {ICLR} 2015, San Diego, CA, USA, May 7-9, 2015, Conference Track Proceedings}}, \bibfield{editor}{\bibinfo{person}{Yoshua Bengio} {and} \bibinfo{person}{Yann LeCun}} (Eds.).
\newblock
\urldef\tempurl%
\url{http://arxiv.org/abs/1409.1556}
\showURL{%
\tempurl}


\bibitem[Singh et~al\mbox{.}(2022)]%
        {DBLP:conf/cvpr/SinghHGCGRK22}
\bibfield{author}{\bibinfo{person}{Amanpreet Singh}, \bibinfo{person}{Ronghang Hu}, \bibinfo{person}{Vedanuj Goswami}, \bibinfo{person}{Guillaume Couairon}, \bibinfo{person}{Wojciech Galuba}, \bibinfo{person}{Marcus Rohrbach}, {and} \bibinfo{person}{Douwe Kiela}.} \bibinfo{year}{2022}\natexlab{}.
\newblock \showarticletitle{{FLAVA:} {A} Foundational Language And Vision Alignment Model}. In \bibinfo{booktitle}{\emph{{IEEE/CVF} Conference on Computer Vision and Pattern Recognition, {CVPR} 2022, New Orleans, LA, USA, June 18-24, 2022}}. \bibinfo{publisher}{{IEEE}}, \bibinfo{pages}{15617--15629}.
\newblock
\urldef\tempurl%
\url{https://doi.org/10.1109/CVPR52688.2022.01519}
\showDOI{\tempurl}


\bibitem[Srivastava et~al\mbox{.}(2014)]%
        {DBLP:journals/jmlr/SrivastavaHKSS14}
\bibfield{author}{\bibinfo{person}{Nitish Srivastava}, \bibinfo{person}{Geoffrey~E. Hinton}, \bibinfo{person}{Alex Krizhevsky}, \bibinfo{person}{Ilya Sutskever}, {and} \bibinfo{person}{Ruslan Salakhutdinov}.} \bibinfo{year}{2014}\natexlab{}.
\newblock \showarticletitle{Dropout: a simple way to prevent neural networks from overfitting}.
\newblock \bibinfo{journal}{\emph{J. Mach. Learn. Res.}} \bibinfo{volume}{15}, \bibinfo{number}{1} (\bibinfo{year}{2014}), \bibinfo{pages}{1929--1958}.
\newblock
\urldef\tempurl%
\url{https://doi.org/10.5555/2627435.2670313}
\showDOI{\tempurl}


\bibitem[Su et~al\mbox{.}(2021)]%
        {DBLP:journals/isci/SuCJZLGWL21}
\bibfield{author}{\bibinfo{person}{Jinsong Su}, \bibinfo{person}{Jinchang Chen}, \bibinfo{person}{Hui Jiang}, \bibinfo{person}{Chulun Zhou}, \bibinfo{person}{Huan Lin}, \bibinfo{person}{Yubin Ge}, \bibinfo{person}{Qingqiang Wu}, {and} \bibinfo{person}{Yongxuan Lai}.} \bibinfo{year}{2021}\natexlab{}.
\newblock \showarticletitle{Multi-modal neural machine translation with deep semantic interactions}.
\newblock \bibinfo{journal}{\emph{Inf. Sci.}}  \bibinfo{volume}{554} (\bibinfo{year}{2021}), \bibinfo{pages}{47--60}.
\newblock
\urldef\tempurl%
\url{https://doi.org/10.1016/j.ins.2020.11.024}
\showDOI{\tempurl}


\bibitem[Sun et~al\mbox{.}(2020)]%
        {DBLP:conf/coling/SunWSWSZC20}
\bibfield{author}{\bibinfo{person}{Lin Sun}, \bibinfo{person}{Jiquan Wang}, \bibinfo{person}{Yindu Su}, \bibinfo{person}{Fangsheng Weng}, \bibinfo{person}{Yuxuan Sun}, \bibinfo{person}{Zengwei Zheng}, {and} \bibinfo{person}{Yuanyi Chen}.} \bibinfo{year}{2020}\natexlab{}.
\newblock \showarticletitle{{RIVA:} {A} Pre-trained Tweet Multimodal Model Based on Text-image Relation for Multimodal {NER}}. In \bibinfo{booktitle}{\emph{Proceedings of the 28th International Conference on Computational Linguistics, {COLING} 2020, Barcelona, Spain (Online), December 8-13, 2020}}. \bibinfo{publisher}{International Committee on Computational Linguistics}, \bibinfo{pages}{1852--1862}.
\newblock
\urldef\tempurl%
\url{https://doi.org/10.18653/v1/2020.coling-main.168}
\showDOI{\tempurl}


\bibitem[Sun et~al\mbox{.}(2021)]%
        {DBLP:conf/aaai/0006W0SW21}
\bibfield{author}{\bibinfo{person}{Lin Sun}, \bibinfo{person}{Jiquan Wang}, \bibinfo{person}{Kai Zhang}, \bibinfo{person}{Yindu Su}, {and} \bibinfo{person}{Fangsheng Weng}.} \bibinfo{year}{2021}\natexlab{}.
\newblock \showarticletitle{RpBERT: {A} Text-image Relation Propagation-based {BERT} Model for Multimodal {NER}}. In \bibinfo{booktitle}{\emph{Thirty-Fifth {AAAI} Conference on Artificial Intelligence, {AAAI} 2021, Thirty-Third Conference on Innovative Applications of Artificial Intelligence, {IAAI} 2021, The Eleventh Symposium on Educational Advances in Artificial Intelligence, {EAAI} 2021, Virtual Event, February 2-9, 2021}}. \bibinfo{publisher}{{AAAI} Press}, \bibinfo{pages}{13860--13868}.
\newblock
\urldef\tempurl%
\url{https://ojs.aaai.org/index.php/AAAI/article/view/17633}
\showURL{%
\tempurl}


\bibitem[Vaswani et~al\mbox{.}(2017)]%
        {DBLP:conf/nips/VaswaniSPUJGKP17}
\bibfield{author}{\bibinfo{person}{Ashish Vaswani}, \bibinfo{person}{Noam Shazeer}, \bibinfo{person}{Niki Parmar}, \bibinfo{person}{Jakob Uszkoreit}, \bibinfo{person}{Llion Jones}, \bibinfo{person}{Aidan~N. Gomez}, \bibinfo{person}{Lukasz Kaiser}, {and} \bibinfo{person}{Illia Polosukhin}.} \bibinfo{year}{2017}\natexlab{}.
\newblock \showarticletitle{Attention is All you Need}. In \bibinfo{booktitle}{\emph{Advances in Neural Information Processing Systems 30: Annual Conference on Neural Information Processing Systems 2017, December 4-9, 2017, Long Beach, CA, {USA}}}. \bibinfo{pages}{5998--6008}.
\newblock
\urldef\tempurl%
\url{https://proceedings.neurips.cc/paper/2017/hash/3f5ee243547dee91fbd053c1c4a845aa-Abstract.html}
\showURL{%
\tempurl}


\bibitem[Wang et~al\mbox{.}(2019)]%
        {DBLP:conf/acl/WangLCKLS19}
\bibfield{author}{\bibinfo{person}{Yue Wang}, \bibinfo{person}{Jing Li}, \bibinfo{person}{Hou~Pong Chan}, \bibinfo{person}{Irwin King}, \bibinfo{person}{Michael~R. Lyu}, {and} \bibinfo{person}{Shuming Shi}.} \bibinfo{year}{2019}\natexlab{}.
\newblock \showarticletitle{Topic-Aware Neural Keyphrase Generation for Social Media Language}. In \bibinfo{booktitle}{\emph{Proceedings of the 57th Conference of the Association for Computational Linguistics, {ACL} 2019, Florence, Italy, July 28- August 2, 2019, Volume 1: Long Papers}}, \bibfield{editor}{\bibinfo{person}{Anna Korhonen}, \bibinfo{person}{David~R. Traum}, {and} \bibinfo{person}{Llu{\'{\i}}s M{\`{a}}rquez}} (Eds.). \bibinfo{publisher}{Association for Computational Linguistics}, \bibinfo{pages}{2516--2526}.
\newblock
\urldef\tempurl%
\url{https://doi.org/10.18653/v1/p19-1240}
\showDOI{\tempurl}


\bibitem[Wang et~al\mbox{.}(2020)]%
        {DBLP:conf/emnlp/WangLLK20}
\bibfield{author}{\bibinfo{person}{Yue Wang}, \bibinfo{person}{Jing Li}, \bibinfo{person}{Michael~R. Lyu}, {and} \bibinfo{person}{Irwin King}.} \bibinfo{year}{2020}\natexlab{}.
\newblock \showarticletitle{Cross-Media Keyphrase Prediction: {A} Unified Framework with Multi-Modality Multi-Head Attention and Image Wordings}. In \bibinfo{booktitle}{\emph{Proceedings of the 2020 Conference on Empirical Methods in Natural Language Processing, {EMNLP} 2020, Online, November 16-20, 2020}}. \bibinfo{publisher}{Association for Computational Linguistics}, \bibinfo{pages}{3311--3324}.
\newblock
\urldef\tempurl%
\url{https://doi.org/10.18653/v1/2020.emnlp-main.268}
\showDOI{\tempurl}


\bibitem[Xie et~al\mbox{.}(2023)]%
        {DBLP:journals/ipm/XieSSWWYLXS23}
\bibfield{author}{\bibinfo{person}{Binbin Xie}, \bibinfo{person}{Jia Song}, \bibinfo{person}{Liangying Shao}, \bibinfo{person}{Suhang Wu}, \bibinfo{person}{Xiangpeng Wei}, \bibinfo{person}{Baosong Yang}, \bibinfo{person}{Huan Lin}, \bibinfo{person}{Jun Xie}, {and} \bibinfo{person}{Jinsong Su}.} \bibinfo{year}{2023}\natexlab{}.
\newblock \showarticletitle{From statistical methods to deep learning, automatic keyphrase prediction: {A} survey}.
\newblock \bibinfo{journal}{\emph{Inf. Process. Manag.}} \bibinfo{volume}{60}, \bibinfo{number}{4} (\bibinfo{year}{2023}), \bibinfo{pages}{103382}.
\newblock
\urldef\tempurl%
\url{https://doi.org/10.1016/j.ipm.2023.103382}
\showDOI{\tempurl}


\bibitem[Xie et~al\mbox{.}(2022)]%
        {DBLP:conf/emnlp/XieWYLXWZS22}
\bibfield{author}{\bibinfo{person}{Binbin Xie}, \bibinfo{person}{Xiangpeng Wei}, \bibinfo{person}{Baosong Yang}, \bibinfo{person}{Huan Lin}, \bibinfo{person}{Jun Xie}, \bibinfo{person}{Xiaoli Wang}, \bibinfo{person}{Min Zhang}, {and} \bibinfo{person}{Jinsong Su}.} \bibinfo{year}{2022}\natexlab{}.
\newblock \showarticletitle{WR-One2Set: Towards Well-Calibrated Keyphrase Generation}. In \bibinfo{booktitle}{\emph{Proceedings of the 2022 Conference on Empirical Methods in Natural Language Processing, {EMNLP} 2022, Abu Dhabi, United Arab Emirates, December 7-11, 2022}}, \bibfield{editor}{\bibinfo{person}{Yoav Goldberg}, \bibinfo{person}{Zornitsa Kozareva}, {and} \bibinfo{person}{Yue Zhang}} (Eds.). \bibinfo{publisher}{Association for Computational Linguistics}, \bibinfo{pages}{7283--7293}.
\newblock
\urldef\tempurl%
\url{https://aclanthology.org/2022.emnlp-main.491}
\showURL{%
\tempurl}


\bibitem[Ye et~al\mbox{.}(2021)]%
        {DBLP:conf/acl/YeGL0Z20}
\bibfield{author}{\bibinfo{person}{Jiacheng Ye}, \bibinfo{person}{Tao Gui}, \bibinfo{person}{Yichao Luo}, \bibinfo{person}{Yige Xu}, {and} \bibinfo{person}{Qi Zhang}.} \bibinfo{year}{2021}\natexlab{}.
\newblock \showarticletitle{One2Set: Generating Diverse Keyphrases as a Set}. In \bibinfo{booktitle}{\emph{Proceedings of the 59th Annual Meeting of the Association for Computational Linguistics and the 11th International Joint Conference on Natural Language Processing, {ACL/IJCNLP} 2021, (Volume 1: Long Papers), Virtual Event, August 1-6, 2021}}. \bibinfo{publisher}{Association for Computational Linguistics}, \bibinfo{pages}{4598--4608}.
\newblock
\urldef\tempurl%
\url{https://doi.org/10.18653/v1/2021.acl-long.354}
\showDOI{\tempurl}


\bibitem[Ye et~al\mbox{.}(2022)]%
        {DBLP:conf/coling/YeGXT022}
\bibfield{author}{\bibinfo{person}{Junjie Ye}, \bibinfo{person}{Junjun Guo}, \bibinfo{person}{Yan Xiang}, \bibinfo{person}{Kaiwen Tan}, {and} \bibinfo{person}{Zhengtao Yu}.} \bibinfo{year}{2022}\natexlab{}.
\newblock \showarticletitle{Noise-robust Cross-modal Interactive Learning with Text2Image Mask for Multi-modal Neural Machine Translation}. In \bibinfo{booktitle}{\emph{Proceedings of the 29th International Conference on Computational Linguistics, {COLING} 2022, Gyeongju, Republic of Korea, October 12-17, 2022}}. \bibinfo{publisher}{International Committee on Computational Linguistics}, \bibinfo{pages}{5098--5108}.
\newblock
\urldef\tempurl%
\url{https://aclanthology.org/2022.coling-1.452}
\showURL{%
\tempurl}


\bibitem[Yin et~al\mbox{.}(2020)]%
        {DBLP:conf/acl/YinMSZYZL20}
\bibfield{author}{\bibinfo{person}{Yongjing Yin}, \bibinfo{person}{Fandong Meng}, \bibinfo{person}{Jinsong Su}, \bibinfo{person}{Chulun Zhou}, \bibinfo{person}{Zhengyuan Yang}, \bibinfo{person}{Jie Zhou}, {and} \bibinfo{person}{Jiebo Luo}.} \bibinfo{year}{2020}\natexlab{}.
\newblock \showarticletitle{A Novel Graph-based Multi-modal Fusion Encoder for Neural Machine Translation}. In \bibinfo{booktitle}{\emph{Proceedings of the 58th Annual Meeting of the Association for Computational Linguistics, {ACL} 2020, Online, July 5-10, 2020}}, \bibfield{editor}{\bibinfo{person}{Dan Jurafsky}, \bibinfo{person}{Joyce Chai}, \bibinfo{person}{Natalie Schluter}, {and} \bibinfo{person}{Joel~R. Tetreault}} (Eds.). \bibinfo{publisher}{Association for Computational Linguistics}, \bibinfo{pages}{3025--3035}.
\newblock
\urldef\tempurl%
\url{https://doi.org/10.18653/v1/2020.acl-main.273}
\showDOI{\tempurl}


\bibitem[Yu et~al\mbox{.}(2022)]%
        {DBLP:conf/ijcai/YuWXL22}
\bibfield{author}{\bibinfo{person}{Jianfei Yu}, \bibinfo{person}{Jieming Wang}, \bibinfo{person}{Rui Xia}, {and} \bibinfo{person}{Junjie Li}.} \bibinfo{year}{2022}\natexlab{}.
\newblock \showarticletitle{Targeted Multimodal Sentiment Classification based on Coarse-to-Fine Grained Image-Target Matching}. In \bibinfo{booktitle}{\emph{Proceedings of the Thirty-First International Joint Conference on Artificial Intelligence, {IJCAI} 2022, Vienna, Austria, 23-29 July 2022}}, \bibfield{editor}{\bibinfo{person}{Luc~De Raedt}} (Ed.). \bibinfo{publisher}{ijcai.org}, \bibinfo{pages}{4482--4488}.
\newblock
\urldef\tempurl%
\url{https://doi.org/10.24963/ijcai.2022/622}
\showDOI{\tempurl}


\bibitem[Yuan et~al\mbox{.}(2020)]%
        {DBLP:conf/acl/YuanWMTBHT20}
\bibfield{author}{\bibinfo{person}{Xingdi Yuan}, \bibinfo{person}{Tong Wang}, \bibinfo{person}{Rui Meng}, \bibinfo{person}{Khushboo Thaker}, \bibinfo{person}{Peter Brusilovsky}, \bibinfo{person}{Daqing He}, {and} \bibinfo{person}{Adam Trischler}.} \bibinfo{year}{2020}\natexlab{}.
\newblock \showarticletitle{One Size Does Not Fit All: Generating and Evaluating Variable Number of Keyphrases}. In \bibinfo{booktitle}{\emph{Proceedings of the 58th Annual Meeting of the Association for Computational Linguistics, {ACL} 2020, Online, July 5-10, 2020}}. \bibinfo{publisher}{Association for Computational Linguistics}, \bibinfo{pages}{7961--7975}.
\newblock
\urldef\tempurl%
\url{https://doi.org/10.18653/v1/2020.acl-main.710}
\showDOI{\tempurl}


\bibitem[Zeng et~al\mbox{.}(2018)]%
        {DBLP:conf/emnlp/ZengLSGLK18}
\bibfield{author}{\bibinfo{person}{Jichuan Zeng}, \bibinfo{person}{Jing Li}, \bibinfo{person}{Yan Song}, \bibinfo{person}{Cuiyun Gao}, \bibinfo{person}{Michael~R. Lyu}, {and} \bibinfo{person}{Irwin King}.} \bibinfo{year}{2018}\natexlab{}.
\newblock \showarticletitle{Topic Memory Networks for Short Text Classification}. In \bibinfo{booktitle}{\emph{Proceedings of the 2018 Conference on Empirical Methods in Natural Language Processing, Brussels, Belgium, October 31 - November 4, 2018}}. \bibinfo{publisher}{Association for Computational Linguistics}, \bibinfo{pages}{3120--3131}.
\newblock
\urldef\tempurl%
\url{https://doi.org/10.18653/v1/d18-1351}
\showDOI{\tempurl}


\bibitem[Zhang et~al\mbox{.}(2020)]%
        {DBLP:journals/jstsp/ZhangYHD20}
\bibfield{author}{\bibinfo{person}{Chao Zhang}, \bibinfo{person}{Zichao Yang}, \bibinfo{person}{Xiaodong He}, {and} \bibinfo{person}{Li Deng}.} \bibinfo{year}{2020}\natexlab{}.
\newblock \showarticletitle{Multimodal Intelligence: Representation Learning, Information Fusion, and Applications}.
\newblock \bibinfo{journal}{\emph{{IEEE} J. Sel. Top. Signal Process.}} \bibinfo{volume}{14}, \bibinfo{number}{3} (\bibinfo{year}{2020}), \bibinfo{pages}{478--493}.
\newblock
\urldef\tempurl%
\url{https://doi.org/10.1109/JSTSP.2020.2987728}
\showDOI{\tempurl}


\bibitem[Zhang et~al\mbox{.}(2017)]%
        {DBLP:conf/ijcai/ZhangWHHG17}
\bibfield{author}{\bibinfo{person}{Qi Zhang}, \bibinfo{person}{Jiawen Wang}, \bibinfo{person}{Haoran Huang}, \bibinfo{person}{Xuanjing Huang}, {and} \bibinfo{person}{Yeyun Gong}.} \bibinfo{year}{2017}\natexlab{}.
\newblock \showarticletitle{Hashtag Recommendation for Multimodal Microblog Using Co-Attention Network}. In \bibinfo{booktitle}{\emph{Proceedings of the Twenty-Sixth International Joint Conference on Artificial Intelligence, {IJCAI} 2017, Melbourne, Australia, August 19-25, 2017}}, \bibfield{editor}{\bibinfo{person}{Carles Sierra}} (Ed.). \bibinfo{publisher}{ijcai.org}, \bibinfo{pages}{3420--3426}.
\newblock
\urldef\tempurl%
\url{https://doi.org/10.24963/ijcai.2017/478}
\showDOI{\tempurl}


\bibitem[Zhang et~al\mbox{.}(2019)]%
        {DBLP:conf/aaai/Zhang00TY019}
\bibfield{author}{\bibinfo{person}{Suwei Zhang}, \bibinfo{person}{Yuan Yao}, \bibinfo{person}{Feng Xu}, \bibinfo{person}{Hanghang Tong}, \bibinfo{person}{Xiaohui Yan}, {and} \bibinfo{person}{Jian Lu}.} \bibinfo{year}{2019}\natexlab{}.
\newblock \showarticletitle{Hashtag Recommendation for Photo Sharing Services}. In \bibinfo{booktitle}{\emph{The Thirty-Third {AAAI} Conference on Artificial Intelligence, {AAAI} 2019, Honolulu, Hawaii, USA, January 27 - February 1, 2019}}. \bibinfo{publisher}{{AAAI} Press}, \bibinfo{pages}{5805--5812}.
\newblock
\urldef\tempurl%
\url{https://doi.org/10.1609/aaai.v33i01.33015805}
\showDOI{\tempurl}


\bibitem[Zhao et~al\mbox{.}(2021)]%
        {DBLP:conf/ner/ZhaoLL21}
\bibfield{author}{\bibinfo{person}{Zhi{-}Wei Zhao}, \bibinfo{person}{Wei Liu}, {and} \bibinfo{person}{Bao{-}Liang Lu}.} \bibinfo{year}{2021}\natexlab{}.
\newblock \showarticletitle{Multimodal Emotion Recognition Using a Modified Dense Co-Attention Symmetric Network}. In \bibinfo{booktitle}{\emph{10th International {IEEE/EMBS} Conference on Neural Engineering, {NER} 2021, Virtual Event, Italy, May 4-6, 2021}}. \bibinfo{publisher}{{IEEE}}, \bibinfo{pages}{73--76}.
\newblock
\urldef\tempurl%
\url{https://doi.org/10.1109/NER49283.2021.9441352}
\showDOI{\tempurl}


\end{thebibliography}
\balance

\newpage
\appendix

\section{Dataset Statistics}
In our experiments, we use two datasets. 

One is the TRC dataset\footnote{https://github.com/danielpreotiuc/text-image-relationship/}, which is used to train the model via the image-text matching task. The dataset contains four types of twitters with different text-image relations. As shown in Table \ref{tab:trc}, according to the roles of images in the text-image pairs, we can roughly divide these pairs into relevant pairs ($R_1\cup R_2$) and irrelevant ones ($R_3\cup R_4$). 

The other is the dataset for multi-modal keyphrase generation collected by \citet{DBLP:conf/emnlp/WangLLK20}.
This dataset includes 53,701 English tweets, each of which comprises a distinct text-image pair, with user-annotated hashtags serving as keyphrases. Table \ref{tab:data} illustrates the dataset statistics.

\begin{table}[]
\caption{Four relation types in the TRC dataset.}
  \label{tab:trc}
\begin{tabular}{lcccc}
\toprule
 & $R_1$  &$R_2$  &$R_3$  &$R_4$  \\
 \midrule
Image adds to the tweet meaning &$\checkmark$  &$\checkmark$  &$\times$    &$\times$     \\
Text is presented in image &$\checkmark$  &$\times$  &$\checkmark$  &$\times$    \\
Percentage (\%) & 18.5 & 25.6 & 21.9 & 33.9 \\
\bottomrule
\end{tabular}
\end{table}

\begin{table}
\caption{Data split statistics.  Avg len: the average token number. KP: keyphrases. |KP|: the number of unique keyphrases. \% of abs.KP: percentage of absent keyphrases. }
  \label{tab:data}
\begin{tabular}{ccccccc}
\toprule
Split & \begin{tabular}[c]{@{}c@{}}\#Post\end{tabular} & \begin{tabular}[c]{@{}c@{}}Avg len \\ per post\end{tabular} & \begin{tabular}[c]{@{}c@{}}\#KP \\ /Post\end{tabular} & |KP| & \begin{tabular}[c]{@{}c@{}}Avg len\\ per KP\end{tabular} & \begin{tabular}[c]{@{}c@{}}\% of\\ abs.KP\end{tabular} \\
\midrule
Train & 42,959 & 27.26 & 1.33 & 4,261 & 1.85 & 37.14 \\
Valid & 5,370 & 26.81 & 1.34 & 2,544 & 1.85 & 36.01 \\
Test & 5,372 & 27.05 & 1.32 & 2,534 & 1.86 & 37.45 \\
\bottomrule
\end{tabular}
\end{table}

\end{document}